\documentclass{article}

\usepackage{microtype}
\usepackage{graphicx}
\graphicspath{{figures/}}
\usepackage{subcaption}
\usepackage{booktabs}
\usepackage{hyperref}

\usepackage[accepted]{icml2026}
\makeatletter
\renewcommand{\ICML@appearing}{\textit{Foundations of Deep
Generative Models: Understanding Memorization, Generalization, and
Reasoning} Workshop at the $\mathit{43}^{rd}$ International
Conference on Machine Learning, Seoul, South Korea, 2026.
Copyright 2026 by the author(s).}
\makeatother

\usepackage{amsmath}
\usepackage{longtable}
\usepackage{amssymb}
\usepackage{mathtools}
\usepackage{amsthm}
\usepackage[capitalize,noabbrev]{cleveref}

\newcommand{\doi}[1]{doi:\ \texttt{\detokenize{#1}}}

\newcommand{\dn}{D/N}
\newcommand{\fstar}{f^{*}}
\newcommand{\cm}{\mathrm{CM}}
\newcommand{\support}{f}
\newcommand{\predpass}{\checkmark}
\newcommand{\predfail}{$\times$}
\newcommand{\predunsc}{$\circ$}

\icmltitlerunning{Natural Ungrokking: Asymmetric Control of Which Rules Survive Pretraining}

\begin{document}

\twocolumn[
\icmltitle{Natural Ungrokking: \\Asymmetric Control of Which Rules Survive Pretraining}

\begin{icmlauthorlist}
\icmlauthor{Juliana Li}{harvard}
\icmlauthor{Diya Sreedhar}{harvard}
\end{icmlauthorlist}
\icmlaffiliation{harvard}{Harvard University, Cambridge, MA, USA}
\icmlcorrespondingauthor{Juliana Li}{julianali@college.harvard.edu}
\icmlkeywords{pretraining dynamics, language models, grokking, natural ungrokking, capability retention, transient capabilities, generalization dynamics, mechanistic interpretability, phase transitions}

\vskip 0.3in
]

\printAffiliationsAndNotice{}

\begin{abstract}
Midway through an ordinary pretraining run, a small language model
learns the pronoun-gender rule: cued with a girl's name
(``\emph{Sue cried because}''), it resolves the next pronoun to
\emph{she}, generalizing to held-out probes ($0.94$ by step 925).
By step 3,500 the same model scores near zero on the same probes,
although the rule's evidence is still in the training data. We
call this within-run reversal \emph{natural ungrokking}: the
corpus decides, with no trace in the loss curve, which learned
rules a model keeps.

Which rules survive is predictable from one corpus statistic: how
often the training stream shows the rule winning. Across
un-intervened runs (two corpora, three data budgets, three
seeds), support frequency decides a rule's fate; the
data-to-parameter ratio only modulates how deeply a doomed rule
falls. The same emerge-then-collapse training dynamics appear in
public Pythia checkpoints, collapse depth ordered by model scale
as predicted. The forgetting is a displacement: a competing
surface pattern out-competes the rule, and the log-probability
margin between them crosses zero within 100 training steps of the
behavioral collapse.

Control over this fate is asymmetric: the same edit that destroys a
rule on demand cannot restore it. Flipping support to
counter-evidence in place kills the rule with monotone dose-response
in two unrelated rules; but injecting support back---even to
$450\times$ the level that naturally sustains it---buys no recovery.
Every confirmatory threshold and prediction was pre-registered
before the data it governed was read.
\end{abstract}

\section{Introduction}
\label{sec:intro}

\begin{figure*}[t]
\centering
\includegraphics[width=\textwidth]{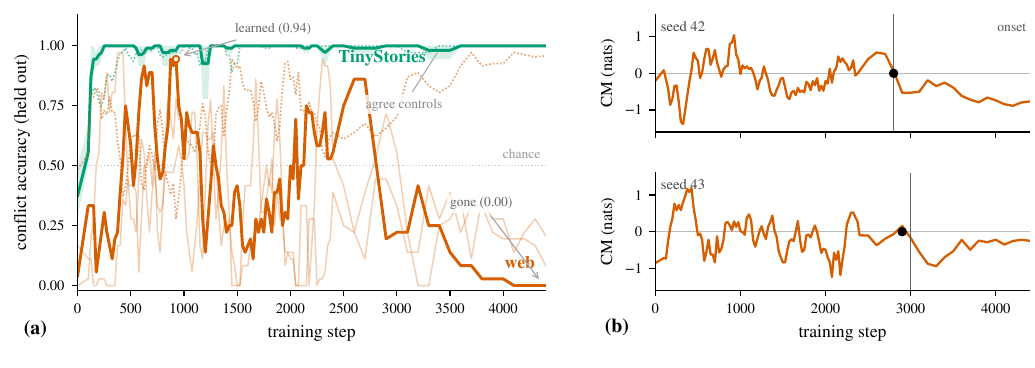}
\caption{The focal pronoun-gender rule emerges and then collapses
under web pretraining, and an internal margin crosses zero as it
does. \textbf{(a)} Held-out conflict accuracy. On TinyStories
(green: seed mean, min--max band) the rule survives at ceiling; on
web the bold vermillion trace is the seed the abstract quotes
($0.94$ by step 925, gone by the end), with the two faint traces
rising and collapsing the same way. Dotted lines are each corpus's
agree-condition control, which keeps climbing through the collapse:
the construction stays solved and only the rule is lost. Grey
dotted marks chance. \textbf{(b)} Contrast margin $\cm$
(\cref{sec:results-mech}), the model's preference for the rule over
the surface default, one strip per instrument-valid web seed; the
vertical line marks behavioral collapse onset, the dot the $\cm$
zero crossing. The two coincide in both seeds (steps
$2{,}800/2{,}800$ and $2{,}900/3{,}000$). The third web seed fails
the instrument check (\cref{sec:setup}) and is omitted; its
crossing is directionally consistent. TinyStories margins stay off
scale at $+2$ to $+12$ nats.}
\label{fig:traj}
\end{figure*}

Within a single pretraining run, a language model can lose a
capability it had already acquired. The case this paper dissects is
the pronoun-gender rule in an
11.5M-parameter model training on web text: cued with a girl's
name, the following pronoun should resolve to \emph{she}. A
conflict probe item is the prefix ``\emph{Sue cried because}'',
scored on the continuation \emph{she} against \emph{he}; the
corpus-wide prior favors \emph{he}, the name cue demands \emph{she}.
Its agree-condition counterpart, ``\emph{Tom cried because}'',
aligns cue and prior and acts as the control. The model scores
$0.94$ on held-out conflict items by step 925 and near $0.00$ by
steps 3{,}500--4{,}400 of the same run, while still scoring $1.00$
when rule and prior agree and the rule's evidence stays in the
stationary stream. No training data was removed, and no distribution
shifted. The model acquired the rule and then
stopped applying it (\cref{fig:traj}a). We call this within-run
reversal \emph{natural ungrokking}.

A capability sighted at a
mid-training checkpoint is no guarantee it will exist in the final
model, but its support frequency in the corpus is predictive of
whether it survives. Data filtering that thins a rare rule's
support can doom that rule without deleting a single example of
it. Continual pretraining on a shifted mix can silently ungrok
capabilities the base model had. In every case the failure is
invisible to loss curves, because the model keeps the construction
and abandons only the rule.

This paper makes three contributions.

\textbf{Capabilities are cheap to destroy and hard to restore.}
Editing a rule's evidence in place, flipping it to counter-evidence
at a chosen rate with token counts and every other corpus statistic
held fixed, destroys the rule with strictly monotone dose-response.
A second, unrelated rule (\emph{a}/\emph{an} allomorphy) replicates
the kill on a five-dose ladder built inside one corpus,
dialing final accuracy monotonically from $0.96$ to $0.00$ while
unrelated capabilities hold at baseline; in both rules, the scored
blind predictions about kill direction all came out correct. The
same knob fails in reverse—we injected support
back into the collapsed corpus at up to $3\times$ the TinyStories
support density, reaching post-injection rule:prior ratios far
above the surviving TinyStories cell; the mechanistic margin
moved, but no run produced a behavioral recovery that
passed its controls (\cref{sec:results-causal}).

\textbf{A rule's fate follows a support-frequency law.}
Whether a rule survives to the end of training is decided by how
often its supporting evidence appears in the corpus; the
unique-data-to-parameter ratio $\dn$ modulates how deeply a doomed
rule falls but never flips its fate (\cref{sec:results-phase}). We
establish this across a grid of corpora, data budgets, and seeds.
The same rise-and-fall reproduces in the smaller public Pythia
checkpoints, with collapse depth following the predicted scale
order, and the verdicts transfer to an out-of-distribution probe
set.

\textbf{The loss is displacement.} The collapsing rule loses a
competition with a strengthening surface pattern, here the
corpus-wide preference for \emph{he}, while the construction it
lives in stays solved. A contrast
margin between rule and prior crosses zero at the behavioral
collapse step, and its final sign largely separates recovered from
displaced cells across the grid, with a boundary case discussed
below (\cref{sec:results-mech}).

Every threshold, metric, and directional prediction here was
registered in advance (\cref{sec:setup}), with the full scoreboard
in App.~\ref{app:prereg}. Predicting a rule's fate works; predicting
\emph{where} on the frequency axis the survival boundary falls is the
open problem this paper hands to future work (App.~\ref{app:theory}).
Code, configs,
probe batteries, and the frozen registration document are
released.\footnote{\url{https://github.com/lijuliana/Natural-Ungrokking}}

\section{Related Work}
\label{sec:related}

Grokking is delayed generalization \citep{power2022grokking,
nanda2023progress}, with phase diagrams drawn over data size and
hyperparameters \citep{liu2022omnigrok, liu2022understanding}. The
nearest dynamical relative of our subject is \emph{ungrokking}
\citep{varma2023explaining}, where generalization recedes when the
dataset shrinks below a critical size between runs; our
capabilities rise and fall within a single run on stationary text,
which is the kinship and the difference that \emph{natural
ungrokking} names. Transience itself has precedent:
\citet{singh2023transient} show emergent in-context learning can
fade under continued training; we observe the analogous fate for
in-weights linguistic rules, give it a frequency law, and steer it
with corpus edits. \citet{chen2024sudden} document sudden
transitions in syntax acquisition and control them through the
training data; we follow the same observational-to-causal arc for
a transition that runs from competence to incompetence.
\citet{chang2024characterizing} show that token-level forgetting
during ordinary pretraining is frequency-dependent; our subject is
a rule-level capability displaced while its evidence stays in the
stream. \citet{wei2021frequency} causally tie agreement behavior
to evidence frequency across separately trained models; we chart
within-run fate. Closest in method is filtered-corpus training
\citep{patil2024fict, misra2024aanns}, which withholds a
construction's direct evidence and finds models acquire it anyway;
our edit instead converts evidence into counter-evidence at fixed
token counts, and the near-parity dose nearest a pure ablation does
not kill the rule, consistent with their results. Extended
discussion is in App.~\ref{app:related-ext}.

\section{Experimental Setup}
\label{sec:setup}

\paragraph{Models and training.}
All governed runs use a 4-layer decoder-only transformer
\citep{vaswani2017attention} with
$d_{\text{model}}=256$, two attention heads, a 2048-token context,
and a corpus-specific BPE vocabulary \citep{sennrich2016bpe} of
8192 symbols (11.5M
parameters; 3.1M excluding embeddings); the architecture and
initialization follow nanochat \citep{karpathy2025nanochat}. Each run trains for
4{,}400 steps at batch size 32 with cosine warmdown over the final
30\% of steps. Learning rates were fixed once per corpus and never
revisited; within any comparison, every cell shares the same
architecture, tokenizer, schedule, and optimizer, so that only the
named axis moves. Each cell is trained with three seeds, and a
disjoint seed set is quarantined for a blind replication pass
(App.~\ref{app:seeds}).

\paragraph{Corpora and the two axes.}
We train on two corpora: TinyStories \citep{eldan2023tinystories},
where the rule under study is densely supported, and a filtered web
corpus derived from ClimbMix \citep{diao2025climb}, where its support
falls below our measurement floor under the registered windowed
counter. The
\emph{support frequency} $\support$ of a rule is the rate of
supporting events per token under a frozen counting procedure
(App.~\ref{app:technical}): intuitively, how often the training
stream shows the rule winning; for the focal rule, an occurrence of
\emph{she} within $16$ tokens of a feminine cue. We write
$\delta_{\mathrm{TS}} = 1.67\times10^{-3}$ events per token for the
focal pronoun-gender rule's measured support frequency on
TinyStories, the unit in which rescue doses are quoted. The
\emph{data ratio} $\dn$ is unique training
tokens over parameters, varied by capping the token budget
($\dn \in \{1.5, 5\}$, plus uncapped full-corpus runs, hereafter
``packed'' cells). The grid spans
both corpora at all budgets, three seeds per cell.

\paragraph{Probe battery and outcome classifier.}
Capabilities are scored by a frozen battery of templated probe
families, each with two conditions: a \emph{conflict} condition,
where the rule's cue and a competing surface cue disagree, and an
\emph{agree} control condition, where they align. A frozen
classifier maps each family's smoothed held-out conflict
trajectory to one of five outcomes (\textsc{recovered},
\textsc{displaced}, \textsc{partial}, \textsc{never},
\textsc{unstable}). A family's verdict counts only when its agree
condition stays solved, which rules out trajectories where the
model has simply lost the construction wholesale: a
\textsc{displaced} verdict means the model still commands the
construction but no longer follows the rule. Thresholds and
smoothing are reproduced verbatim in App.~\ref{app:technical}.

\paragraph{Mechanism metric.}
Alongside behavior we track a contrast margin $\cm$: on a frozen
prompt set disjoint from every training intervention, the mean
log-probability margin for the rule-conforming continuation over
its prior-conforming competitor. For the focal rule, on
feminine-cue prompts $x$,
\begin{equation}
\cm \;=\; \tfrac{1}{|\mathcal{P}|}\textstyle\sum_{x \in \mathcal{P}}
\bigl[\log p_\theta(\textit{she} \mid x) -
\log p_\theta(\textit{he} \mid x)\bigr],
\label{eq:cm}
\end{equation}
so $\cm > 0$ means the rule outweighs the prior at the prediction
site and $\cm < 0$ means the prior has won
(App.~\ref{app:technical}). An instrument guard discards runs
where smoothed $\cm$ never attains $0.5$ nats. $\cm$ is computed
independently of the behavioral classifier, so mechanism and
behavior can disagree, and in places they do.

\paragraph{Pre-registration protocol.}
Every confirmatory metric, threshold, and directional prediction was
committed to version control before the data it governs was read;
later changes are dated amendments, failed predictions are reported
as failures, and post-hoc analyses are marked as such and carry no
scored verdicts (App.~\ref{app:prereg}). Two validity gates recur
and leave a prediction \emph{unscoreable} rather than passed or
failed: an \emph{instrument} gate when a metric fails its own
preconditions (e.g.\ the $0.5$-nat margin guard), and an
\emph{intervention} gate when an edited corpus damages unrelated
control families.

\section{Results}
\label{sec:results}

The survival pattern
(\cref{sec:results-phase}) establishes the law, the mechanism
(\cref{sec:results-mech}) shows what the loss is, and the causal
interventions (\cref{sec:results-causal}) test it in both
directions. Every confirmatory claim below is a registered
prediction, scored on the complete pass/fail scoreboard in
App.~\ref{app:prereg} (\cref{tab:predictions}).

\subsection{Support Frequency Determines Rule Fate at Every Data
Budget}
\label{sec:results-phase}

\begin{figure}[tb]
\centering
\includegraphics[width=\columnwidth]{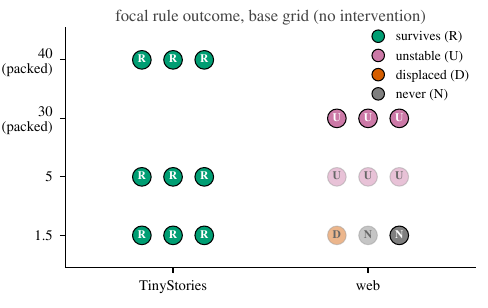}
\caption{Base grid for the focal pronoun-gender rule: one marker
per seed per cell, lettered by the frozen classifier's verdict
(\textsc{R}ecovered, \textsc{U}nstable, \textsc{D}isplaced,
\textsc{N}ever); faded markers are seeds whose agree control failed
(unscoreable). Rows are the unique-data/params budgets (ordinal
spacing); the packed row has corpus-specific $\dn$, approximately
$30$ for TinyStories and $40$ for web. The rule survives in every
TinyStories cell and no web cell; the budget axis never flips the
verdict.}
\label{fig:phasegrid}
\end{figure}

All four blind predictions made with the grid design (P1--P4)
passed. \Cref{tab:phase} (App.~\ref{app:seeds}) gives the per-seed
outcomes for all six families and \cref{fig:phasegrid} plots the
focal rule's grid. The focal pronoun-gender rule survives (ends
\textsc{recovered} under the frozen classifier) in 9/9 TinyStories
runs at every budget, and in no web run at any budget: on the packed
web cells it emerges, collapses, and is pinned at the competing
prior. The one web run that approaches the threshold
($\dn{=}5$, seed 42, conflict final $0.770$ against the $0.8$
survival bar) is a boundary case that returns in
\cref{sec:results-mech}. All four high-support families keep conflict final $\geq 0.8$ on
packed web in 3/3 seeds (P2's registered criterion, which scores
conflict finals alone): two are fully \textsc{recovered} across the
grid, and the other two solve their conflict items but carry
unscoreable agree-control verdicts from a probe-calibration artifact.
The \texttt{reflexive\_gender} family ends below threshold on web,
and the web budget ordering holds. Across all cells, $\dn$ modulates
displacement depth but never flips a survival verdict; the fate
axis is support frequency.

Two control cells rule out deflationary explanations. An
optimizer-control cell (TinyStories under the web cell's optimizer)
ends at conflict $0.927$, so optimizer settings alone cannot cause
the collapse; and noun-cued and name-cued items both end below the
survival threshold on web, ruling out forgetting of particular name
embeddings. The loss sits at the level of the rule.

The grid realizes its support-frequency axis through corpus
identity, so the two corpora differ in many statistics at once.
Clean attribution to support frequency therefore comes from the
interventions of \cref{sec:results-causal}, which edit that one
statistic in place, including a five-dose ladder within a single
corpus where the same law reappears.

\begin{figure}[t]
\centering
\includegraphics[width=\columnwidth]{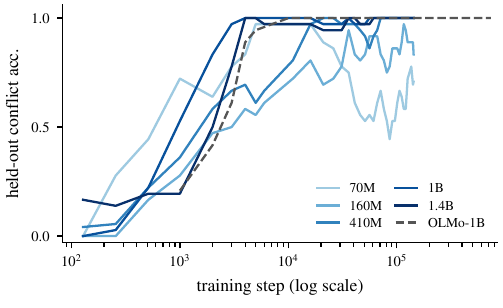}
\caption{The transient signature in public checkpoints, same frozen
battery: held-out conflict accuracy for the focal rule across Pythia
revisions (70M--1.4B) and OLMo-1B, frozen $k{=}3$ smoothing,
log-axis steps. The rule emerges early in every model;
end-of-training survival is ordered by scale (PUB3, Spearman
$\rho = 0.894$), the smallest model falling deepest and collapse
gone by 410M. Agree controls of the collapsing models stay at or
near ceiling throughout (not shown).}
\label{fig:public}
\end{figure}

\paragraph{Public-checkpoint transfer.} The phenomenon is not an
artifact of our 11.5M-parameter setting (\cref{fig:public}). Scored
with the same battery, the focal rule emerges early in every public
model (Pythia 70M--1.4B and OLMo-1B, trained by other labs on other
corpora), then collapses in the smaller ones while the agree control
stays solved. The predicted depth-by-scale ordering transfers to
models two orders of magnitude larger: the smaller the model, the
deeper the final collapse (Spearman $\rho = 0.894$ across five Pythia
sizes, collapse gone by 410M). Exact magnitudes did not: one
prediction failed by $0.008$ and a boundary census missed
(App.~\ref{app:prereg}).

\subsection{A Contrast Margin Crosses Zero as Behavior Collapses}
\label{sec:results-mech}

Two hypotheses compete when the rule disappears:
\emph{erasure} (the construction is lost wholesale) and
\emph{displacement} (the rule loses a race with a strengthening
surface default fed far more often, so the model keeps the
construction but stops honoring the rule where the two disagree).
Two controls already constrain the answer: collapsed cells keep
solving the construction when rule and prior agree (against
erasure), and on web cells train-template finals track held-out
finals within $0.11$ (against pure memorization).

The displacement signature lives in the contrast margin $\cm$
(\cref{sec:setup}), and mean final $\cm$ is a clean order parameter
for the phase diagram: $+3.68 / +2.98 / +2.36$ across TinyStories
budgets (packed, $\dn{=}5$, $\dn{=}1.5$) against
$-0.52 / -0.07 / -0.85$ across the same web budgets, with the
run-level sign rule (final $\cm$ positive exactly when final
conflict accuracy exceeds $0.5$) holding in 18/18 base runs. The
same scalar moves with the causal knob, the kill driving $\cm$ from
$+3.68$ to $-2.99$ dose-monotonically (\cref{sec:results-causal}),
exactly as displacement predicts.

A registered prediction ties the margin to behavior in time. In
both web seeds where the margin instrument is valid, the smoothed
$\cm$ zero-crossing lands within one $100$-step checkpoint of the
behavioral collapse onset (\cref{fig:traj}b): crossings at steps
$2{,}800$ vs.\ $2{,}800$ and $2{,}900$ vs.\ $3{,}000$ (the third
seed fails the guard; its crossing is directionally consistent).
The margin crosses \emph{as} behavior collapses, not after, and
every crossing precedes the cosine warmdown (step $3{,}080$ of
$4{,}400$), ruling out a schedule artifact.

A second, gate-free measure corroborates this on all three web
seeds (post-hoc, descriptive; App.~\ref{app:mech2}). An exact
decomposition of the \emph{she}--\emph{he} logit gap into a direct
embedding-path term and a contextual term (all attention and MLP
contributions) shows the direct term staying near zero while the
contextual term carries the margin and ends negative: the collapse
is in the contextual pathway, not the static readout. A
parameter-free probe agrees, cue-gender decodability at the
prediction site decaying toward chance on web while staying near
ceiling on surviving TinyStories runs: the displacement reaches down
into the representation, upstream of the logits.

Head-level analyses (exploratory; App.~\ref{app:mech2},
methods follow \citealp{elhage2021mathematical}) support this picture: 
in surviving runs the margin rides a single last-layer head
($75$--$90\%$ of attribution, zero-ablation deletes it), while
collapsed runs have no dominant carrier and no fitted gender
direction moves the margin by more than $0.22$ nats. Patching within
a single run reproduces the same destroy-easily/restore-hardly
asymmetry at the circuit level (App.~\ref{app:mech2}).

One registered margin verdict failed (App.~\ref{app:prereg}): an
all-cells sign prediction assumed every web run ends below the $0.5$
sign threshold, but the boundary run of \cref{sec:results-phase}
ended just above it (conflict final $0.770$, still under the $0.8$
survival bar), triggering the mechanism falsifier on that universal
claim. The timing result stands on its own test: wherever the margin
instrument is healthy, the margin crosses zero just as behavior dies.

\subsection{Causal Control: An Asymmetry Between Removing and
Restoring Support}
\label{sec:results-causal}

\begin{figure*}[t]
\centering
\includegraphics[width=\textwidth]{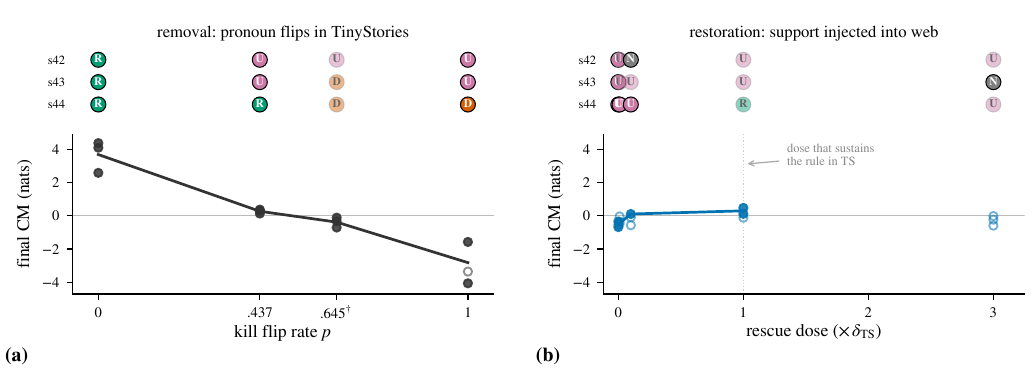}
\caption{The two intervention directions, each on its own dose axis
with the un-intervened base at dose zero. Top strips: per-seed
behavioral outcome (colors as in \cref{fig:phasegrid}); faded
markers failed a registered validity gate (their control condition,
or in the $\dag$ cell the whole-cell intervention check). Bottom:
final $\cm$, one point per seed; hollow points fail the $\cm$
instrument gate, the line is the mean over gate-passing seeds.
Removal \textbf{(a)} is monotone: more flipped support, deeper
collapse. Restoration \textbf{(b)} never produces a control-valid
recovery (none even at three times the sustaining dose) and shifts
the margin only weakly, peaking at matched dose and regressing
beyond it. $\dag$: the intervention-invalid kill cell
(\cref{tab:step6}).}
\label{fig:phase}
\end{figure*}

The causal experiment's result illuminates an asymmetry:
the same one-variable edit that destroys a rule on demand fails,
run in reverse at up to triple strength, to bring the rule back.
The survival pattern and the mechanism both nominate support
frequency as the controlling variable; the causal experiment edits
it in place. In the
surviving TinyStories cell, we flip the rule-supporting pronoun
token after girl-name cues at rates
$p \in \{0.437, 0.645, 1.0\}$, chosen in advance to bring the
rule-to-prior support ratio to parity, to match the web corpus's
ratio of $0.46$, and to remove all support, leaving the corpus
byte-identical elsewhere with token counts unchanged. The ladder
spans the withheld-evidence regime of filtered-corpus training
\citep{patil2024fict, misra2024aanns} and, at $p{=}1$, a strictly
stronger contradiction regime. Blind directional predictions and
per-cell validity gates preceded any intervened corpus
(App.~\ref{app:prereg}). \Cref{fig:phase} plots the results; the
per-cell numbers are in \cref{tab:step6} (App.~\ref{app:rescue}).

\paragraph{Removal is dose-graded and predictable.} At full flip
the rule dies in 3/3 seeds with $\cm$ collapsing from $+3.68$
(base) to $-2.99$, while the near-parity dose $p{=}0.437$ does not
kill the cell ($0/3$ seeds \textsc{displaced}, mean margin
$+0.27$). Monotonicity is exact: $\rho_{\mathrm{kill}} = -1.00$
over the four-point ladder of base plus three doses. With four
points a perfect ordering has a one-sided null probability of
$1/24$, so the evidential weight rests on the blind registration
of direction, threshold, and falsifier, all of which came out as
predicted. The middle dose damaged two
unrelated control families in 2/3 seeds, so its behavioral verdict
is voided by the intervention gate; scored at face value it would
also have been correct (3/3 died, $\cm$ $-0.38$), and excluding
the cell entirely the remaining points stay strictly monotone.
The lethal boundary lies between parity and full flip, and
\emph{every scored prediction in the removal direction was
correct} (3/3: full-flip behavior, full-flip margin, dose
monotonicity).

\paragraph{A second rule, one corpus.} The kill generalizes, in
the setting that matters most for the grid's confound
(\cref{sec:results-phase}): support varied at five doses
\emph{within} a single corpus, so corpus identity cannot carry the
effect. The rule is \emph{a}/\emph{an} allomorphy (choose \emph{an}
before a vowel-initial word), at ceiling in base TinyStories with a
rule:counter evidence ratio of $23.4$. We flip \emph{an} to
\emph{a} before vowel-initial words at rates
$p \in \{0.5, 0.667, 0.75, 0.9, 1.0\}$, three seeds each, every
other statistic untouched, with five blind predictions made in
advance (App.~\ref{app:ankill}). Final conflict accuracy falls
strictly monotonically in dose (\cref{tab:ankill}), cell means
$0.96 \to 0.67 \to 0.49 \to 0.29 \to 0.13 \to 0.00$ as the post-edit
evidence ratio falls $23.4 \to 0.92 \to 0.47 \to 0.32 \to 0.11 \to 0$. The edit is
surgical: pronoun-gender, negation, and irregular-past families
stay within $0.05$ of base at every dose, while a second probe
family of the same rule (\emph{a}/\emph{an} before adjectives)
co-degrades, $0.99 \to 0.00$. The intermediate doses reproduce the
grid's phase structure inside one corpus: at $p{=}0.667$ the rule
emerges and then collapses; at $p{=}1.0$ it never emerges. Two
predictions missed (App.~\ref{app:ankill}): the rule destabilizes
already at the parity dose we predicted it would survive, so both
rules' boundaries sit within a factor of two of evidence parity, and
the agree-condition control we predicted would fail at high dose
stayed valid everywhere.

\begin{table}[t]
\caption{The \emph{a}/\emph{an} kill ladder, run entirely within one
TinyStories corpus: per-seed and mean final held-out conflict
accuracy (frozen-classifier letter R/U/N) as the post-edit
rule:counter evidence ratio falls. The rule degrades strictly
monotonically to zero, and the adjacency family
\texttt{a\_an\_adj.} co-degrades while unrelated families hold
(App.~\ref{app:ankill}).}
\label{tab:ankill}
\centering\small
\resizebox{0.98\columnwidth}{!}{
\begin{tabular}{lrlllrr}
\toprule
 & & \multicolumn{4}{c}{\texttt{det\_an\_choice}} & \texttt{a\_an*} \\
\cmidrule(lr){3-6}
flip rate $p$ & r:c & \multicolumn{3}{c}{final (letter), seeds 42/43/44} & mean & mean \\
\midrule
0 (base) & 23.4 & $0.95$\,(R) & $0.98$\,(R) & $0.96$\,(R) & $0.96$ & $0.99$ \\
0.5 & 0.92 & $0.76$\,(U) & $0.69$\,(U) & $0.55$\,(U) & $0.67$ & $0.74$ \\
0.667 & 0.47 & $0.66$\,(U) & $0.37$\,(N) & $0.45$\,(U) & $0.49$ & $0.64$ \\
0.75 & 0.32 & $0.28$\,(U) & $0.13$\,(U) & $0.47$\,(U) & $0.29$ & $0.35$ \\
0.9 & 0.11 & $0.25$\,(U) & $0.07$\,(U) & $0.08$\,(N) & $0.13$ & $0.03$ \\
1 & 0.00 & $0.00$\,(N) & $0.00$\,(N) & $0.00$\,(N) & $0.00$ & $0.00$ \\
\bottomrule
\end{tabular}
}
\end{table}

\paragraph{Restoration fails at matched and overshot dose.} The
restoration arm injects the same kind of support the kill removes
(\emph{girl}$\to$\emph{she} documents with symmetric \emph{boy}$\to$\emph{he} controls, token
counts matched) into the collapsed web corpus at doses up to
$3\times\delta_{\mathrm{TS}}$, where $1\times\delta_{\mathrm{TS}}$
is the support density under which the rule survives $3/3$ seeds
on TinyStories. No dose produces a control-valid recovery in any
seed, tripping the failure condition we committed to in advance for
this arm. The margin registers a partial effect: at the
$1\times$ dose final $\cm$ turns positive as predicted (cell mean
$+0.16$) but lands an order of magnitude below the $+2.4$ to
$+3.7$ of the surviving regime, and behavior does not follow. The
moved margin is visible in the circuit too: at this dose the dominant
last-layer head measurably re-forms, with roughly double the
ablation cost of the un-injected base, yet the item-level
confidence interval on the behavioral drop still excludes zero in
every non-artifact-flagged rescue run (App.~\ref{app:mech2}): the
dose produces a partial reassembly of the circuit while the behavior
stays collapsed.

A unit check closes a natural objection: that density-matched
doses might still be ratio-poor, the injected evidence drowned by
the corpus prior. The opposite holds: under the same windowed
counter, run post hoc on the injected cue class
(App.~\ref{app:rescue}), the post-injection rule:prior ratio is
already $37$ at the smallest dose and $3{,}565$ at $1\times$
(\cref{tab:step6}), against the $7.9$ of the TinyStories cell
where the rule survives. By the evidence-ratio standard that
governs every other cell in this paper, the injected construction
is over-supported by orders of magnitude, and behavior still does
not recover. The injected names are deliberately disjoint from the
battery's cues (an anti-teaching control), so recovery requires the
class-level rule to generalize across names; a model that merely
memorized the injected sentences would not register on the battery.
Within this design, support frequency is causally sufficient for
destruction and insufficient for restoration. One objection remained
for the uniform-rate design: about a third of the injected dose arrives
only after the collapse step, so the failure could have reflected a
mistimed dose. A registered amendment (dated after the
uniform-rate verdicts were read, before any timing run existed)
concentrated the full $1\times$ dose into the window before
the collapse step, with a matched late-window control. The early dose also failed: zero of
three seeds show a control-valid recovery, and the one seed that
held the rule while the dose flowed lost it once the dose ended.
Full verdicts and the instrument gates that leave the mechanism
comparison unscored are in App.~\ref{app:rescue}.

\paragraph{Out-of-distribution transfer.} All verdicts above are
on the templated battery; an OOD battery (864 held-out items,
frames disjoint from all training injections,
App.~\ref{app:technical}) rules out template artifacts. All three
transfer predictions passed, competence, failure, and the transient
emerge-then-collapse signature, and the kill dose gradient and
collapsed restoration cells both reproduce out of distribution.

\section{Discussion and Limitations}
\label{sec:discussion}

A learned capability was cheap to destroy, with exact
dose-response in two unrelated rules, and could not be bought back
by re-adding matched evidence: the same knob, run in reverse at up
to triple strength, moved the internal margin modestly, with
partial re-formation of the margin-carrying head, while restoring
none of the behavior. The strongest registered objection was dose
timing, and the asymmetry survived it: concentrating the full dose
before the collapse step still produced zero control-valid
recoveries (\cref{sec:results-causal}). At every dose and schedule
we tested, matched data failed to buy back what pretraining had
discarded, a pattern consistent with the selection of survivors
being made early and then consolidating.

The other two results say when to expect the loss and what it
physically is. Natural ungrokking is predictable: a rule's
end-of-training fate can be read off a measurable corpus
statistic, its support frequency, before training ends, and
nothing about it is adversarial: the stream is stationary, and a
rare rule's support is simply outweighed by a more frequent
competing pattern. And the loss takes a
physical form, displacement, measurable as a margin-level order
parameter that crosses zero exactly when behavior collapses. The
practical consequences named in \cref{sec:intro} follow from these
three results jointly: mid-training evaluation, data filtering,
and continual pretraining all interact with a survival law that
is invisible to loss curves.

Several limitations bound the scope of these claims.
\textbf{Scale:} the intervention is established at 11.5M
parameters; the public suite shows the phenomenon and its
scale-ordering to 1.4B, but no intervention has been run at larger
scale. \textbf{Probes:} all primary verdicts use templated minimal
pairs; the OOD battery transfers all three of its predictions and a
surface $n$-gram baseline cannot reproduce the phase pattern
(App.~\ref{app:mech-alt}), though both batteries are synthetic and
naturalistic evaluation is left for future work.
\textbf{The asymmetry:} its mechanism comparison rests on the one
early-window seed with a valid margin readout, and its behavioral
verdict spans three seeds per window and a single rule, so
consolidation past a point of no return, which fits the results,
remains to be measured directly (App.~\ref{app:rescue}).
\textbf{Replication:} the quarantined seed set and its registered
predictions remain unread by design (App.~\ref{app:seeds}).

What we hand to future work is \emph{where} on the frequency axis
the boundary falls. We registered a critical-frequency theory of
exactly that before inspecting any counts; it predicted which rules
survive but not the location of the boundary, so its quantitative
form is left open (post-mortem in App.~\ref{app:theory}), and the
one candidate that fits the grid awaits the blind replication pass.
The phenomenon that theory would complete already stands on firm
ground: a rule's fate is predictable from a corpus statistic before
the run ends, the loss is displacement at the mechanism level, and
that fate is controllable in one direction by a single knob, all
registered in advance.

\section*{Impact Statement}
This work studies how language-model capabilities form and
disappear during pretraining, and its practical thrust is
constructive: a capability a model shows mid-run can be anticipated,
monitored, and to a degree controlled before the run ends. If the
findings generalize, three consequences for practice follow. A
capability observed at a mid-training checkpoint can be absent from
the final model, so evaluation timing belongs in release decisions.
Corpus filtering can remove a rare rule without deleting a single
example of it. Continual pretraining on a shifted mix can quietly
undo rules a base model had already learned. None of the three
shows up in the loss curve, which argues for capability-level
monitoring during training rather than only after it.

Two responsibilities attend the method. The kill intervention edits
gendered-pronoun statistics in a small synthetic corpus only to test
a causal claim about rule retention at 11.5M parameters; the same
technique aimed at capability removal at scale would carry dual-use
concerns, and we report it so that defenders can recognize and
monitor for it. The probes measure a grammatical resolution rule,
whether a name cue overrides a frequency prior, so no conclusion
about social bias in deployed models follows from them. Finally, the
pre-registration protocol we follow, frozen thresholds and blind
directional predictions with passes and failures reported alike, is
itself a contribution we hope lowers the rate of
unscrutinized claims in training-dynamics research.

\bibliography{fogen}

\begin{thebibliography}{48}
\providecommand{\natexlab}[1]{#1}
\providecommand{\url}[1]{\texttt{#1}}
\expandafter\ifx\csname urlstyle\endcsname\relax
  \providecommand{\doi}[1]{doi: #1}\else
  \providecommand{\doi}{doi: \begingroup \urlstyle{rm}\Url}\fi

\bibitem[Achille et~al.(2019)Achille, Rovere, and Soatto]{achille2019critical}
Achille, A., Rovere, M., and Soatto, S.
\newblock Critical learning periods in deep networks.
\newblock In \emph{International Conference on Learning Representations}, 2019.
\newblock arXiv:1711.08856.

\bibitem[Biderman et~al.(2023)Biderman, Schoelkopf, Anthony, Bradley, O'Brien, Hallahan, Khan, Purohit, Prashanth, Raff, Skowron, Sutawika, and van~der Wal]{biderman2023pythia}
Biderman, S., Schoelkopf, H., Anthony, Q., Bradley, H., O'Brien, K., Hallahan, E., Khan, M.~A., Purohit, S., Prashanth, U.~S., Raff, E., Skowron, A., Sutawika, L., and van~der Wal, O.
\newblock Pythia: A suite for analyzing large language models across training and scaling.
\newblock In \emph{International Conference on Machine Learning}, 2023.
\newblock arXiv:2304.01373.

\bibitem[Brants et~al.(2007)Brants, Popat, Xu, Och, and Dean]{brants2007large}
Brants, T., Popat, A.~C., Xu, P., Och, F.~J., and Dean, J.
\newblock Large language models in machine translation.
\newblock In \emph{Proceedings of the 2007 Joint Conference on Empirical Methods in Natural Language Processing and Computational Natural Language Learning (EMNLP-CoNLL)}, pp.\  858--867, 2007.

\bibitem[Chang \& Bergen(2022)Chang and Bergen]{chang2022word}
Chang, T.~A. and Bergen, B.~K.
\newblock Word acquisition in neural language models.
\newblock \emph{Transactions of the Association for Computational Linguistics}, 10:\penalty0 1--16, 2022.
\newblock \doi{10.1162/tacl_a_00444}.

\bibitem[Chang et~al.(2024)Chang, Tu, and Bergen]{chang2024characterizing}
Chang, T.~A., Tu, Z., and Bergen, B.~K.
\newblock Characterizing learning curves during language model pre-training: Learning, forgetting, and stability.
\newblock \emph{Transactions of the Association for Computational Linguistics}, 12:\penalty0 1346--1362, 2024.
\newblock \doi{10.1162/tacl_a_00708}.

\bibitem[Chen et~al.(2024)Chen, Shwartz-Ziv, Cho, Leavitt, and Saphra]{chen2024sudden}
Chen, A., Shwartz-Ziv, R., Cho, K., Leavitt, M.~L., and Saphra, N.
\newblock Sudden drops in the loss: Syntax acquisition, phase transitions, and simplicity bias in mlms.
\newblock In \emph{International Conference on Learning Representations}, 2024.
\newblock arXiv:2309.07311.

\bibitem[Choshen et~al.(2022)Choshen, Hacohen, Weinshall, and Abend]{choshen2022grammar}
Choshen, L., Hacohen, G., Weinshall, D., and Abend, O.
\newblock The grammar-learning trajectories of neural language models.
\newblock In \emph{Proceedings of the 60th Annual Meeting of the Association for Computational Linguistics}, 2022.

\bibitem[Diao et~al.(2025)Diao, Yang, Fu, Dong, Su, Kliegl, Chen, Belcak, Suhara, Yin, Patwary, Lin, Kautz, and Molchanov]{diao2025climb}
Diao, S., Yang, Y., Fu, Y., Dong, X., Su, D., Kliegl, M., Chen, Z., Belcak, P., Suhara, Y., Yin, H., Patwary, M., Lin, Y., Kautz, J., and Molchanov, P.
\newblock Nemotron-climb: Clustering-based iterative data mixture bootstrapping for language model pre-training.
\newblock \emph{arXiv preprint arXiv:2504.13161}, 2025.

\bibitem[Eldan \& Li(2023)Eldan and Li]{eldan2023tinystories}
Eldan, R. and Li, Y.
\newblock Tinystories: How small can language models be and still speak coherent english?
\newblock \emph{arXiv preprint arXiv:2305.07759}, 2023.

\bibitem[Elhage et~al.(2021)Elhage, Nanda, Olsson, Henighan, Joseph, Mann, Askell, Bai, Chen, Conerly, et~al.]{elhage2021mathematical}
Elhage, N., Nanda, N., Olsson, C., Henighan, T., Joseph, N., Mann, B., Askell, A., Bai, Y., Chen, A., Conerly, T., et~al.
\newblock A mathematical framework for transformer circuits.
\newblock \emph{Transformer Circuits Thread}, 2021.
\newblock \url{https://transformer-circuits.pub/2021/framework/index.html}.

\bibitem[{Gemma Team} et~al.(2024){Gemma Team}, Rivi{\`e}re, Pathak, Sessa, Hardin, Bhupatiraju, Hussenot, Mesnard, Shahriari, Ram{\'e}, et~al.]{gemmateam2024gemma2}
{Gemma Team}, Rivi{\`e}re, M., Pathak, S., Sessa, P.~G., Hardin, C., Bhupatiraju, S., Hussenot, L., Mesnard, T., Shahriari, B., Ram{\'e}, A., et~al.
\newblock Gemma 2: Improving open language models at a practical size.
\newblock \emph{arXiv preprint arXiv:2408.00118}, 2024.

\bibitem[Groeneveld et~al.(2024)Groeneveld, Beltagy, Walsh, Bhagia, Kinney, Tafjord, Jha, Ivison, Magnusson, Wang, et~al.]{groeneveld2024olmo}
Groeneveld, D., Beltagy, I., Walsh, P., Bhagia, A., Kinney, R., Tafjord, O., Jha, A.~H., Ivison, H., Magnusson, I., Wang, Y., et~al.
\newblock Olmo: Accelerating the science of language models.
\newblock In \emph{Proceedings of the 62nd Annual Meeting of the Association for Computational Linguistics}, 2024.
\newblock arXiv:2402.00838.

\bibitem[Hoffmann et~al.(2022)Hoffmann, Borgeaud, Mensch, Buchatskaya, Cai, Rutherford, de~Las~Casas, Hendricks, Welbl, Clark, et~al.]{hoffmann2022chinchilla}
Hoffmann, J., Borgeaud, S., Mensch, A., Buchatskaya, E., Cai, T., Rutherford, E., de~Las~Casas, D., Hendricks, L.~A., Welbl, J., Clark, A., et~al.
\newblock Training compute-optimal large language models.
\newblock \emph{arXiv preprint arXiv:2203.15556}, 2022.

\bibitem[Hoogland et~al.(2024)Hoogland, Wang, Farrugia-Roberts, Carroll, Wei, and Murfet]{hoogland2024developmental}
Hoogland, J., Wang, G., Farrugia-Roberts, M., Carroll, L., Wei, S., and Murfet, D.
\newblock The developmental landscape of in-context learning.
\newblock \emph{arXiv preprint arXiv:2402.02364}, 2024.

\bibitem[Jagielski et~al.(2022)Jagielski, Thakkar, Tram{\`e}r, Ippolito, Lee, Carlini, Wallace, Song, Thakurta, Papernot, and Zhang]{jagielski2023measuring}
Jagielski, M., Thakkar, O., Tram{\`e}r, F., Ippolito, D., Lee, K., Carlini, N., Wallace, E., Song, S., Thakurta, A., Papernot, N., and Zhang, C.
\newblock Measuring forgetting of memorized training examples.
\newblock \emph{arXiv preprint arXiv:2207.00099}, 2022.

\bibitem[Jordan et~al.(2024)Jordan, Jin, Boza, You, Cesista, Newhouse, and Bernstein]{jordan2024muon}
Jordan, K., Jin, Y., Boza, V., You, J., Cesista, F., Newhouse, L., and Bernstein, J.
\newblock Muon: An optimizer for hidden layers in neural networks.
\newblock \url{https://kellerjordan.github.io/posts/muon/}, 2024.

\bibitem[Kandpal et~al.(2023)Kandpal, Deng, Roberts, Wallace, and Raffel]{kandpal2023longtail}
Kandpal, N., Deng, H., Roberts, A., Wallace, E., and Raffel, C.
\newblock Large language models struggle to learn long-tail knowledge.
\newblock In \emph{International Conference on Machine Learning}, 2023.

\bibitem[Karpathy(2025)]{karpathy2025nanochat}
Karpathy, A.
\newblock nanochat.
\newblock \url{https://github.com/karpathy/nanochat}, 2025.

\bibitem[Kirkpatrick et~al.(2017)Kirkpatrick, Pascanu, Rabinowitz, Veness, Desjardins, Rusu, Milan, Quan, Ramalho, Grabska-Barwinska, Hassabis, Clopath, Kumaran, and Hadsell]{kirkpatrick2017ewc}
Kirkpatrick, J., Pascanu, R., Rabinowitz, N., Veness, J., Desjardins, G., Rusu, A.~A., Milan, K., Quan, J., Ramalho, T., Grabska-Barwinska, A., Hassabis, D., Clopath, C., Kumaran, D., and Hadsell, R.
\newblock Overcoming catastrophic forgetting in neural networks.
\newblock \emph{Proceedings of the National Academy of Sciences}, 114\penalty0 (13):\penalty0 3521--3526, 2017.
\newblock arXiv:1612.00796.

\bibitem[Lindsey et~al.(2024)Lindsey, Templeton, Marcus, Conerly, Batson, and Olah]{lindsey2024crosscoders}
Lindsey, J., Templeton, A., Marcus, J., Conerly, T., Batson, J., and Olah, C.
\newblock Sparse crosscoders for cross-layer features and model diffing.
\newblock Transformer Circuits Thread, 2024.
\newblock URL \url{https://transformer-circuits.pub/2024/crosscoders/index.html}.

\bibitem[Liu et~al.(2022)Liu, Kitouni, Nolte, Michaud, Tegmark, and Williams]{liu2022understanding}
Liu, Z., Kitouni, O., Nolte, N., Michaud, E.~J., Tegmark, M., and Williams, M.
\newblock Towards understanding grokking: An effective theory of representation learning.
\newblock In \emph{Advances in Neural Information Processing Systems}, 2022.

\bibitem[Liu et~al.(2023)Liu, Michaud, and Tegmark]{liu2022omnigrok}
Liu, Z., Michaud, E.~J., and Tegmark, M.
\newblock Omnigrok: Grokking beyond algorithmic data.
\newblock In \emph{International Conference on Learning Representations}, 2023.
\newblock arXiv:2210.01117.

\bibitem[Loshchilov \& Hutter(2019)Loshchilov and Hutter]{loshchilov2019adamw}
Loshchilov, I. and Hutter, F.
\newblock Decoupled weight decay regularization.
\newblock In \emph{International Conference on Learning Representations}, 2019.

\bibitem[Lu et~al.(2020)Lu, Mardziel, Wu, Amancharla, and Datta]{lu2020gender}
Lu, K., Mardziel, P., Wu, F., Amancharla, P., and Datta, A.
\newblock Gender bias in neural natural language processing.
\newblock In \emph{Logic, Language, and Security}. Springer, 2020.

\bibitem[Meng et~al.(2022)Meng, Bau, Andonian, and Belinkov]{meng2022locating}
Meng, K., Bau, D., Andonian, A., and Belinkov, Y.
\newblock Locating and editing factual associations in {GPT}.
\newblock In \emph{Advances in Neural Information Processing Systems}, 2022.

\bibitem[Misra \& Mahowald(2024)Misra and Mahowald]{misra2024aanns}
Misra, K. and Mahowald, K.
\newblock Language models learn rare phenomena from less rare phenomena: The case of the missing {AANNs}.
\newblock \emph{arXiv preprint arXiv:2403.19827}, 2024.

\bibitem[Muennighoff et~al.(2023)Muennighoff, Rush, Barak, Le~Scao, Piktus, Tazi, Pyysalo, Wolf, and Raffel]{muennighoff2023data}
Muennighoff, N., Rush, A.~M., Barak, B., Le~Scao, T., Piktus, A., Tazi, N., Pyysalo, S., Wolf, T., and Raffel, C.
\newblock Scaling data-constrained language models.
\newblock In \emph{Advances in Neural Information Processing Systems}, 2023.
\newblock arXiv:2305.16264.

\bibitem[Murty et~al.(2023)Murty, Sharma, Andreas, and Manning]{murty2023hierarchical}
Murty, S., Sharma, P., Andreas, J., and Manning, C.
\newblock Grokking of hierarchical structure in vanilla transformers.
\newblock In \emph{Proceedings of the 61st Annual Meeting of the Association for Computational Linguistics (Volume 2: Short Papers)}, pp.\  439--448, 2023.
\newblock \doi{10.18653/v1/2023.acl-short.38}.

\bibitem[Nanda et~al.(2023)Nanda, Chan, Lieberum, Smith, and Steinhardt]{nanda2023progress}
Nanda, N., Chan, L., Lieberum, T., Smith, J., and Steinhardt, J.
\newblock Progress measures for grokking via mechanistic interpretability.
\newblock In \emph{International Conference on Learning Representations}, 2023.
\newblock arXiv:2301.05217.

\bibitem[Olsson et~al.(2022)Olsson, Elhage, Nanda, Joseph, DasSarma, Henighan, Mann, Askell, Bai, Chen, et~al.]{olsson2022induction}
Olsson, C., Elhage, N., Nanda, N., Joseph, N., DasSarma, N., Henighan, T., Mann, B., Askell, A., Bai, Y., Chen, A., et~al.
\newblock In-context learning and induction heads.
\newblock \emph{Transformer Circuits Thread}, 2022.
\newblock arXiv:2209.11895.

\bibitem[Patil et~al.(2024)Patil, Jumelet, Chiu, Lapastora, Shen, Wang, Willrich, and Steinert-Threlkeld]{patil2024fict}
Patil, A., Jumelet, J., Chiu, Y.~Y., Lapastora, A., Shen, P., Wang, L., Willrich, C., and Steinert-Threlkeld, S.
\newblock Filtered corpus training ({FiCT}) shows that language models can generalize from indirect evidence.
\newblock \emph{Transactions of the Association for Computational Linguistics}, 12:\penalty0 1597--1615, 2024.
\newblock \doi{10.1162/tacl_a_00720}.

\bibitem[Power et~al.(2022)Power, Burda, Edwards, Babuschkin, and Misra]{power2022grokking}
Power, A., Burda, Y., Edwards, H., Babuschkin, I., and Misra, V.
\newblock Grokking: Generalization beyond overfitting on small algorithmic datasets.
\newblock \emph{arXiv preprint arXiv:2201.02177}, 2022.

\bibitem[Schaeffer et~al.(2023)Schaeffer, Miranda, and Koyejo]{schaeffer2023mirage}
Schaeffer, R., Miranda, B., and Koyejo, S.
\newblock Are emergent abilities of large language models a mirage?
\newblock In \emph{Advances in Neural Information Processing Systems}, 2023.
\newblock arXiv:2304.15004.

\bibitem[Sennrich et~al.(2016)Sennrich, Haddow, and Birch]{sennrich2016bpe}
Sennrich, R., Haddow, B., and Birch, A.
\newblock Neural machine translation of rare words with subword units.
\newblock In \emph{Proceedings of the 54th Annual Meeting of the Association for Computational Linguistics}, 2016.

\bibitem[Singh et~al.(2023)Singh, Chan, Moskovitz, Grant, Saxe, and Hill]{singh2023transient}
Singh, A.~K., Chan, S.~C., Moskovitz, T., Grant, E., Saxe, A.~M., and Hill, F.
\newblock The transient nature of emergent in-context learning in transformers.
\newblock In \emph{Advances in Neural Information Processing Systems}, 2023.
\newblock arXiv:2311.08360.

\bibitem[Tirumala et~al.(2022)Tirumala, Markosyan, Zettlemoyer, and Aghajanyan]{tirumala2022memorization}
Tirumala, K., Markosyan, A.~H., Zettlemoyer, L., and Aghajanyan, A.
\newblock Memorization without overfitting: Analyzing the training dynamics of large language models.
\newblock In \emph{Advances in Neural Information Processing Systems}, 2022.
\newblock arXiv:2205.10770.

\bibitem[Toneva et~al.(2019)Toneva, Sordoni, des Combes, Trischler, Bengio, and Gordon]{toneva2019forgetting}
Toneva, M., Sordoni, A., des Combes, R.~T., Trischler, A., Bengio, Y., and Gordon, G.~J.
\newblock An empirical study of example forgetting during deep neural network learning.
\newblock In \emph{International Conference on Learning Representations}, 2019.
\newblock arXiv:1812.05159.

\bibitem[Varma et~al.(2023)Varma, Shah, Kenton, Kram{\'a}r, and Kumar]{varma2023explaining}
Varma, V., Shah, R., Kenton, Z., Kram{\'a}r, J., and Kumar, R.
\newblock Explaining grokking through circuit efficiency.
\newblock \emph{arXiv preprint arXiv:2309.02390}, 2023.

\bibitem[Vaswani et~al.(2017)Vaswani, Shazeer, Parmar, Uszkoreit, Jones, Gomez, Kaiser, and Polosukhin]{vaswani2017attention}
Vaswani, A., Shazeer, N., Parmar, N., Uszkoreit, J., Jones, L., Gomez, A.~N., Kaiser, {\L}., and Polosukhin, I.
\newblock Attention is all you need.
\newblock In \emph{Advances in Neural Information Processing Systems}, 2017.

\bibitem[Vig et~al.(2020)Vig, Gehrmann, Belinkov, Qian, Nevo, Singer, and Shieber]{vig2020causal}
Vig, J., Gehrmann, S., Belinkov, Y., Qian, S., Nevo, D., Singer, Y., and Shieber, S.
\newblock Investigating gender bias in language models using causal mediation analysis.
\newblock In \emph{Advances in Neural Information Processing Systems}, 2020.

\bibitem[Wang et~al.(2024)Wang, Hoogland, van Wingerden, Furman, and Murfet]{wang2024refined}
Wang, G., Hoogland, J., van Wingerden, S., Furman, Z., and Murfet, D.
\newblock Differentiation and specialization of attention heads via the refined local learning coefficient.
\newblock \emph{arXiv preprint arXiv:2410.02984}, 2024.

\bibitem[Warstadt et~al.(2020{\natexlab{a}})Warstadt, Parrish, Liu, Mohananey, Peng, Wang, and Bowman]{warstadt2020blimp}
Warstadt, A., Parrish, A., Liu, H., Mohananey, A., Peng, W., Wang, S.-F., and Bowman, S.~R.
\newblock Blimp: The benchmark of linguistic minimal pairs for english.
\newblock \emph{Transactions of the Association for Computational Linguistics}, 8:\penalty0 377--392, 2020{\natexlab{a}}.
\newblock arXiv:1912.00582.

\bibitem[Warstadt et~al.(2020{\natexlab{b}})Warstadt, Zhang, Li, Liu, and Bowman]{warstadt2020learning}
Warstadt, A., Zhang, Y., Li, X., Liu, H., and Bowman, S.~R.
\newblock Learning which features matter: {RoBERTa} acquires a preference for linguistic generalizations (eventually).
\newblock In \emph{Proceedings of the 2020 Conference on Empirical Methods in Natural Language Processing}, pp.\  217--235, 2020{\natexlab{b}}.
\newblock \doi{10.18653/v1/2020.emnlp-main.16}.

\bibitem[Wei et~al.(2021)Wei, Garrette, Linzen, and Pavlick]{wei2021frequency}
Wei, J., Garrette, D., Linzen, T., and Pavlick, E.
\newblock Frequency effects on syntactic rule learning in transformers.
\newblock In \emph{Proceedings of the 2021 Conference on Empirical Methods in Natural Language Processing}, pp.\  932--948, 2021.
\newblock \doi{10.18653/v1/2021.emnlp-main.72}.

\bibitem[Wei et~al.(2022)Wei, Tay, Bommasani, Raffel, Zoph, Borgeaud, Yogatama, Bosma, Zhou, Metzler, Chi, Hashimoto, Vinyals, Liang, Dean, and Fedus]{wei2022emergent}
Wei, J., Tay, Y., Bommasani, R., Raffel, C., Zoph, B., Borgeaud, S., Yogatama, D., Bosma, M., Zhou, D., Metzler, D., Chi, E.~H., Hashimoto, T., Vinyals, O., Liang, P., Dean, J., and Fedus, W.
\newblock Emergent abilities of large language models.
\newblock \emph{Transactions on Machine Learning Research}, 2022.
\newblock arXiv:2206.07682.

\bibitem[Xie et~al.(2023)Xie, Pham, Dong, Du, Liu, Lu, Liang, Le, Ma, and Yu]{xie2023doremi}
Xie, S.~M., Pham, H., Dong, X., Du, N., Liu, H., Lu, Y., Liang, P., Le, Q.~V., Ma, T., and Yu, A.~W.
\newblock Doremi: Optimizing data mixtures speeds up language model pretraining.
\newblock In \emph{Advances in Neural Information Processing Systems}, 2023.
\newblock arXiv:2305.10429.

\bibitem[Zhao et~al.(2018)Zhao, Wang, Yatskar, Ordonez, and Chang]{zhao2018gender}
Zhao, J., Wang, T., Yatskar, M., Ordonez, V., and Chang, K.-W.
\newblock Gender bias in coreference resolution: Evaluation and debiasing methods.
\newblock In \emph{Proceedings of NAACL-HLT}, 2018.

\bibitem[Zucchet et~al.(2025)Zucchet, Bornschein, Chan, Lampinen, Pascanu, and De]{zucchet2025facts}
Zucchet, N., Bornschein, J., Chan, S., Lampinen, A., Pascanu, R., and De, S.
\newblock How do language models learn facts? dynamics, curricula and hallucinations.
\newblock \emph{arXiv preprint arXiv:2503.21676}, 2025.

\end{thebibliography}
\bibliographystyle{icml2026}

\newpage
\appendix
\onecolumn

\section{Pre-registration Ledger}
\label{app:prereg}

Every confirmatory metric, threshold, and directional prediction in
this paper was committed to version control \emph{before} the data it
governs existed or was read. The consolidated registration document
(\texttt{prereg/PREREGISTRATION.md}) was frozen at the tag
\texttt{prereg-v1}, with all later changes as dated amendment
sections appended to it. That document and tag, together with the
evaluator and analysis code, are released verbatim
(App.~\ref{app:repro}), so a reader can verify that each
registration preceded the data it governs. \Cref{tab:predictions} summarizes every registered
prediction by group, and \cref{tab:pred-ledger} lists them one by
one. ID prefixes: P grid,
PUB public checkpoints, F frequency theory, S/M mechanism, R rescue,
K kill, RV5 out-of-distribution, CMR replication week (unread).

\begin{table}[ht]
\caption{All registered predictions by group: passed
(\predpass), failed (\predfail), or unscoreable
under an instrument or intervention validity gate
(\predunsc, never converted to \predpass\ or \predfail). The
falsifier column reports each group's registered hard falsifier
(\emph{silent} = untriggered, \emph{triggered} = fired, --- = none);
the summary row counts triggered among the seven applicable
falsifiers (2 of 7). The Step-6T falsifier is silent on a thinner
base than the others: its margin-level clauses are unscoreable
because the timing-window instrument gate passed in only $1/3$
seeds.}
\label{tab:predictions}
\centering\small
\begin{tabular}{@{}l ccc l@{}}
\toprule
prediction group & \predpass & \predfail & \predunsc & falsifier \\
\midrule
Budget-only pilot (pre-grid) & 0 & 1 & 0 & --- \\
Phase grid (Step 3) & 4 & 0 & 0 & silent \\
Public suite (Pythia/OLMo) & 2 & 2 & 0 & --- \\
Critical-frequency theory (Step 4) & 0 & 2 & 0 & silent \\
Mechanism (Step 5) & 3 & 1 & 2 & \textbf{triggered} \\
Causal control (Step 6): rescue & 3 & 3 & 2 & \textbf{triggered} \\
Causal control (Step 6): kill & 3 & 0 & 2 & silent \\
Rescue timing (Step-6T) & 1 & 1 & 2 & silent \\
Second-rule (a/an) kill ladder & 3 & 2 & 0 & --- \\
OOD transfer (rvp5) & 3 & 0 & 0 & silent \\
\midrule
all & 22 & 12 & 8 & 2 of 7 \\
\bottomrule
\end{tabular}

\end{table}

\begin{table}[h]
\caption{Complete per-prediction ledger behind the group summary of
\cref{tab:predictions}: every registered prediction, its verdict
(\predpass\ pass, \predfail\ fail, \predunsc\ unscoreable under a
registered validity gate), and the gate or margin where one applies.
R1's margin clause is displayed as its own row but is a registered
sub-clause of R1, giving the registered Step-6 count of 12
predictions (4 unscoreable; 5 of the 8 scored correct).}
\label{tab:pred-ledger}
\centering\small
\begin{tabular}{@{}l c l@{}}
\toprule
registered prediction & verdict & note \\
\midrule
\multicolumn{3}{@{}l}{\emph{Budget-only pilot (pre-grid)}} \\
\quad $\dn{=}1.5$ displacement predictions & \predfail & falsified \\
\multicolumn{3}{@{}l}{\emph{Phase grid (Step 3) (falsifier silent)}} \\
\quad P1 focal rule: survives all TS, dies web-packed & \predpass &  \\
\quad P2 high-support families survive web & \predpass &  \\
\quad P3 reflexive below 0.8 on web & \predpass &  \\
\quad P4 web budget ordering (packed $\leq$ d15) & \predpass &  \\
\multicolumn{3}{@{}l}{\emph{Public suite (Pythia/OLMo)}} \\
\quad PUB1 boundary-case census & \predfail & decomposed \\
\quad PUB2 pythia-70m displacement depth & \predfail & by 0.008 \\
\quad PUB2$'$ secondary (reflexive) & \predpass &  \\
\quad PUB3 depth vs.\ scale ordering & \predpass &  \\
\multicolumn{3}{@{}l}{\emph{Critical-frequency theory (Step 4) (falsifier silent)}} \\
\quad F1 support-ratio/outcome ordering & \predfail &  \\
\quad F2 displaced families lowest ratio & \predfail &  \\
\multicolumn{3}{@{}l}{\emph{Mechanism (Step 5) (falsifier triggered)}} \\
\quad M1/M2 $S_1$ embedding-geometry scalar & \predunsc & instrument failure \\
\quad M3 no memorized-template residue & \predpass &  \\
\quad M4b $\cm$ zero-cross times collapse & \predpass &  \\
\quad M4a/M4c displacement direction & \predunsc & $<2$ valid seeds \\
\quad M4$'$-TS sign prediction, TS cells & \predpass &  \\
\quad M4$'$-A sign prediction, all cells & \predfail & falsifier hit \\
\multicolumn{3}{@{}l}{\emph{Causal control (Step 6): rescue (falsifier triggered)}} \\
\quad R1 behavioral rescue at $d{\geq}1$ & \predfail &  \\
\quad R1 $\cm$ crosses positive at $d{=}1$ & \predpass &  \\
\quad R1c behavioral rescue at $d{=}3$ & \predfail &  \\
\quad R1c $\cm$ at $d{=}3$ & \predunsc & 0 $\cm$-valid seeds \\
\quad R2 no rescue at trace dose & \predpass &  \\
\quad R2 trace-dose $\cm$ & \predunsc & 0 $\cm$-valid seeds \\
\quad R3 dose monotonicity $\rho\geq0.8$ & \predfail & $\rho=+0.70$ \\
\quad RS rescue specificity & \predpass &  \\
\multicolumn{3}{@{}l}{\emph{Causal control (Step 6): kill (falsifier silent)}} \\
\quad K1 super-critical kill ($p{=}.645$), beh. & \predunsc & KS gate \\
\quad K1 super-critical kill ($p{=}.645$), $\cm$ & \predunsc & KS gate \\
\quad K1c full kill ($p{=}1$), behavioral & \predpass &  \\
\quad K1c full kill ($p{=}1$), $\cm$ & \predpass &  \\
\quad K2 dose monotonicity $\rho\leq-0.8$ & \predpass & $\rho=-1.00$ \\
\multicolumn{3}{@{}l}{\emph{Rescue timing (Step-6T) (falsifier silent)}} \\
\quad T1 early-window dose rescues, beh. & \predfail &  \\
\quad T1 early-window $\cm$ ends positive & \predunsc & G4 gate (1/3 valid) \\
\quad T2 late-window dose fails, beh. & \predpass &  \\
\quad T3 early ${>}$ late $\cm$ ordering & \predunsc & G4 gate \\
\multicolumn{3}{@{}l}{\emph{Second-rule (a/an) kill ladder}} \\
\quad AK1 dose monotonicity (cell means) & \predpass &  \\
\quad AK2 full flip ($p{=}1$) kills & \predpass &  \\
\quad AK3 boundary within $2\times$ of pronoun's & \predfail & crossed lower \\
\quad AK4 agree-gate failure at high dose & \predfail & gate stayed valid \\
\quad AK5 dissociation (unrelated families) & \predpass &  \\
\multicolumn{3}{@{}l}{\emph{OOD transfer (rvp5) (falsifier silent)}} \\
\quad RV5-P1 competence transfers & \predpass &  \\
\quad RV5-P2 failure transfers & \predpass &  \\
\quad RV5-P3 transience transfers & \predpass &  \\
\midrule
Blind directional hit rate (Step 6) & \multicolumn{2}{l}{5/8 scored (12 registered, 4 unscoreable)} \\
\bottomrule
\end{tabular}

\end{table}

\paragraph{Quarantine attestations.} Replication seeds 1042--1044 were
designated before any training at those seeds, trained with identical
configurations, and have never been evaluated, inspected, or plotted;
they are reserved for the registered replication-week predictions
(CMR1/CMR2, commit \texttt{1164c6f}). Quarantined artifacts from failed
builds are retained and renamed, never silently deleted.

\section{The Critical-Frequency Theory: Statement, Verdicts, and
Post-Mortem}
\label{app:theory}

The theory is a critical-frequency account: a rule survives when its
supporting evidence outweighs the competing surface prior it must
override, so outcomes should track a \emph{support ratio}
$(\mathrm{rule\ support}+1)/(\mathrm{prior\ support}+1)$ computed by
a frozen counter over the tokenized training stream
(App.~\ref{app:technical}). We registered it in falsifiable ranking
form before inspecting any counts: (F1) within each web cell,
support ratio and conflict final should correlate positively across
control-valid families; (F2) the displaced gender families should
have the lowest web support ratios; plus a hard falsifier (a cell
whose bottom-two-ratio families recover while its top-two are
displaced).

The qualitative account survives; its registered quantitative form
does not. The hard falsifier was not triggered (no cell shows the
inverted pattern that would kill the theory outright), but both
ranking predictions failed. F1: Spearman by web cell is $+0.800$
(packed) but $-0.500$ (both capped budgets), over the 3--4
control-valid families per cell; with so few points per cell the
coefficients are weakly determined. F2: under the registered windowed
counter the gender families do not have the lowest ratios
(\texttt{a\_an} $2.50$ and \texttt{negation} $4.85$ sit below
pronoun/reflexive at $6.94$). An instrument check found the
registered counter undercounts on web text (the girl-name cue class
tokenizes to zero events; deciding ratios rest on 5--85 counts).
What predicts \emph{where} the boundary lies therefore remains open.
Descriptively, a simple bigram counter puts the two collapsing
gender families' web ratios below $1$ (pronoun $4547/9908 = 0.46$,
reflexive $28/57 = 0.50$), the only sub-unity ratios in either
corpus; the registered test of this stricter two-regime form belongs
to replication week (App.~\ref{app:seeds}), not to this paper's
claims. The $0.46$ kill-dose target of \cref{sec:results-causal} is
the web rule-to-prior ratio under this bigram counter, adopted for
dose design after the windowed counter's blindness post-mortem was
already on file, and frozen together with the rates themselves.

\section{Reproducibility and Compute}
\label{app:repro}

\paragraph{Code, configs, artifacts.} All training and evaluation
code is released at
\url{https://github.com/lijuliana/Natural-Ungrokking}. Configuration
files are the single source of truth for hyperparameters; code reads
configs and never hardcodes them, and every run archives the exact
\texttt{config\_used.yaml} it was launched with. All checkpoints,
frozen probe batteries, eval logs (JSONL), and analysis outputs are
mirrored to a versioned, append-only object store; nothing is ever
deleted, including artifacts of failed or superseded builds (retained
and renamed). Every figure and table in this paper is produced by a
script in the repository reading those artifacts; no number is entered
by hand.

\paragraph{Hardware and wall-clock.} All models in the main grid are
the same 11.5M-parameter architecture (\cref{sec:setup}). One training run
(4{,}400 steps) takes roughly 30 minutes on a single NVIDIA A10G;
training jobs are launched from templated cloud configurations, and
offline battery rescoring runs on CPU or a single A10G. Public-suite
evaluation (Pythia 70M--1.4B, OLMo-1B across training revisions) used
a single A100-80GB for the largest models and A10Gs below 410M.
From the run tracking logs, a training run averages ${\approx}0.43$
A10G-hours: the base grid and registered control cells
(23 runs) total ${\approx}10$ A10G-hours, the Step-6 intervention
cells (21 runs) ${\approx}9$, the Step-6T timing cells (6 runs)
${\approx}3$, the a/an kill ladder (15 runs) ${\approx}6$, and the
quarantined replication seeds
(18 runs, trained but never read) ${\approx}8$, roughly 36
A10G-hours of training in all, plus offline rescoring and the
public-suite evaluation pass.

\paragraph{Frozen evaluators and the independent verifier.} Every
confirmatory read is executed by an evaluator committed before the
governed data was read (App.~\ref{app:prereg}). For the causal-control
step, a second verifier was written from scratch against the
registered text (not the evaluator code), and required to agree with
the frozen evaluator on every per-run field and all 19 verdict keys on
synthesized cells before any real result was read; any future
disagreement is adjudicated against the registered text and logged.

\paragraph{Determinism checks.} Intervened corpora (kill/rescue) are
built deterministically; manifests were rebuilt on independent
machines and verified byte-identical by checksum before training
attached to them, and the registered support-counter instrument check
verified post-intervention support ratios against their predicted
values before launch (App.~\ref{app:prereg}).

\section{Seeds, Robustness, and Replication}
\label{app:seeds}

Every cell in every grid is trained at three exploratory seeds (42,
43, 44). A cell-level claim requires the registered behavior in at
least 2/3 seeds; per-seed outcomes are always reported. A registered
rule forbids adding seeds after a verdict is read; seeds may be
added to a thin cell only by a dated amendment made \emph{before}
reading them, preventing seed-shopping.

\begin{table}[h]
\caption{Per-seed final outcomes (seeds 42/43/44) for six probe
families across the grid. Letters: \texttt{R}~\textsc{recovered},
\texttt{D}~\textsc{displaced}, \texttt{N}~\textsc{never},
\texttt{U}~\textsc{unstable} (\textsc{partial} did not occur in any
run); \textcolor{gray}{grey} letters mark
seeds whose agree (control) condition failed, which yield no
classifiable verdict.}
\label{tab:phase}
\centering
\begin{tabular}{@{}l cccccc@{}}
\toprule
 & \multicolumn{3}{c}{TinyStories ($D/N$)} & \multicolumn{3}{c}{web ($D/N$)} \\
\cmidrule(lr){2-4} \cmidrule(lr){5-7}
family & packed & 5 & 1.5 & packed & 5 & 1.5 \\
\midrule
pronoun\_gender (focal) & \texttt{RRR} & \texttt{RRR} & \texttt{RRR} & \texttt{UUU} & \texttt{\textcolor{gray}{U}\textcolor{gray}{U}\textcolor{gray}{U}} & \texttt{\textcolor{gray}{D}\textcolor{gray}{N}N} \\
reflexive\_gender & \texttt{RRU} & \texttt{RRR} & \texttt{RUR} & \texttt{\textcolor{gray}{U}\textcolor{gray}{U}N} & \texttt{\textcolor{gray}{U}U\textcolor{gray}{N}} & \texttt{\textcolor{gray}{U}\textcolor{gray}{U}N} \\
det\_an\_choice & \texttt{RRR} & \texttt{RRR} & \texttt{RRU} & \texttt{RRR} & \texttt{RRR} & \texttt{RRR} \\
a\_an\_adjective & \texttt{RRR} & \texttt{RRR} & \texttt{\textcolor{gray}{U}RR} & \texttt{RRR} & \texttt{RRR} & \texttt{RRR} \\
irregular\_past & \texttt{RRR} & \texttt{RRR} & \texttt{RRR} & \texttt{\textcolor{gray}{R}\textcolor{gray}{R}\textcolor{gray}{R}} & \texttt{\textcolor{gray}{R}\textcolor{gray}{R}\textcolor{gray}{R}} & \texttt{\textcolor{gray}{R}\textcolor{gray}{R}\textcolor{gray}{R}} \\
negation\_bare\_verb & \texttt{RRR} & \texttt{RRR} & \texttt{RRR} & \texttt{\textcolor{gray}{R}\textcolor{gray}{R}\textcolor{gray}{R}} & \texttt{\textcolor{gray}{R}\textcolor{gray}{R}\textcolor{gray}{R}} & \texttt{\textcolor{gray}{R}\textcolor{gray}{R}\textcolor{gray}{R}} \\
\bottomrule
\end{tabular}

\end{table}

Separately, a quarantined replication seed set (1042--1044) was
designated before any training at those seeds. These runs were trained
with byte-identical configurations but have never been evaluated,
inspected, or plotted. Two replication predictions over them (CMR1/
CMR2, registered at commit \texttt{1164c6f} before any replication
data was read) will be scored exactly once, by the frozen pipeline,
during a designated replication week. At submission time the
quarantine is intact.

\section{Technical Definitions}
\label{app:technical}

\paragraph{Forced-choice scoring.} Each battery item is a (prefix,
correct continuation, distractor) triple with a
\texttt{family.condition} probe id, a template id, and a
train-template/held-out split. The score of a continuation is its
total token log-probability given the prefix; an item is correct iff
the correct continuation's log-probability strictly exceeds the
distractor's (ties score as incorrect). The primary metric is argmax
accuracy aggregated per (probe, split); the secondary metric is the
length-normalized log-probability difference. Classification uses the
held-out split only.

\paragraph{Transience classifier (frozen constants).} Trajectories are
smoothed with a rolling mean of width $k{=}3$ over scored checkpoints
with step $\ge 100$. The final value is the mean of the last 10\% of
evals (at least 3). A family-cell-seed is \emph{emerged} if smoothed
conflict accuracy reaches $\ge 0.8$ at any checkpoint. The class is
assigned in fixed order: \textsc{unstable} if the smoothed conflict
range over the final window is $\ge 0.2$ (still moving at the stop
point, so no final-state label applies); else \textsc{recovered} if
final $\ge 0.8$; else \textsc{displaced} if emerged and final
$\le 0.6$; else \textsc{partial} if emerged (final strictly between
$0.6$ and $0.8$); else \textsc{never} (never emerged). A trajectory
is \emph{control-valid}
iff the agree-condition final is $\ge 0.8$ \emph{and} the agree drop
(max minus final) is $< 0.15$. The class is a final-state property:
dip-then-recover versus dip-then-stuck.

\paragraph{Training details (from the frozen configs).} Muon
\citep{jordan2024muon} on
matrix parameters and AdamW \citep{loshchilov2019adamw}
($\beta = (0.9, 0.95)$) on embeddings
and scalars. TinyStories cells: matrix LR $0.04$, embedding LR
$0.6$, weight decay $0.2$ decayed linearly to zero over training;
web cells: matrix LR $0.02$, embedding LR $0.2$, weight decay $0$.
Batch 32 sequences of $2{,}048$ tokens ($65{,}536$ tokens/step),
$4{,}400$ steps, cosine warmdown over the final $30\%$ (from step
$3{,}080$). Learning rates were fixed per corpus by a registered
sweep before any governed run and never revisited.

\paragraph{Support counter ($\fstar$-v2).} For family $F$ in corpus
$C$, $\mathrm{support\_ratio}(F,C) = (\mathrm{rule\_support} + 1) /
(\mathrm{prior\_support} + 1)$ with exact-pattern counts over the
tokenized training stream. Gender families use windowed mode: an event
is the first `` she''/`` he'' within $k{=}16$ tokens of a single-token
gendered cue, counting both space variants of each cue; other families
use exact bigrams. The TinyStories girl-class support frequency under
this counter is $\delta_{\mathrm{TS}} = 1.67175\times10^{-3}$
events/token, the unit in which
rescue doses are expressed.

\paragraph{Contrast margins (CM/PM) and instrument guard.} On a frozen
prompt set disjoint from all training injections, $\mathrm{PM}$ is the
he-minus-she logit margin on neutral prompts and $\cm$ the she-minus-he
margin on feminine-cue prompts, each smoothed with the same $k{=}3$
convention. A run is CM-instrument-valid iff the maximum smoothed
$\cm$ is $\ge +0.5$ nats at some step $\ge 100$; cells with fewer than
two instrument-valid seeds yield no CM verdict
(\textsc{instrument-invalid}), leaving behavioral verdicts unaffected.
All instruments carry such validity preconditions; violations yield
\textsc{instrument-invalid}, never \textsc{pass}/\textsc{fail}.

\paragraph{Confidence intervals.} Behavioral accuracies carry
template-stratified bootstrap CIs: items are resampled within
template strata, 1{,}000 resamples, fixed RNG seed; CIs are reported
per (probe, split, checkpoint).

\paragraph{Out-of-distribution battery (rvp5).} 864 held-out items in
three families whose sentence frames are disjoint from all prior
batteries and from every rescue-injection frame. Registered before any
scoring: a seed is RVP5-valid iff OOD agree final $\ge 0.7$; transfer
predictions use $\ge 0.7$ (recovered side, a fixed $0.1$ OOD allowance
relative to the templated $0.8$) and $\le 0.6$ (displaced side,
unchanged); the transience-transfer prediction requires OOD smoothed
max $\ge$ final $+\,0.15$ wherever the templated trajectory shows the
transient signature. The registered falsifier: if in two or more
recovered-and-valid cells at least 2/3 of valid seeds score OOD
conflict final $< 0.6$, the transfer claim fails and is reported as a
main-text limitation. In the causal-control cells, per-seed OOD
conflict finals are $0.57/0.72/0.83$ at kill rate $p{=}0.437$,
$0.48/0.16/0.27$ at $0.645$ (descriptive; that cell is
intervention-invalid), and $0.22/0.20/0.05$ at $1.0$; the
restoration cells remain collapsed OOD ($0.08$--$0.33$).

\section{The Restoration (Rescue) Arm in Full}
\label{app:rescue}

This appendix reports every registered verdict of the restoration
arm summarized in \cref{sec:results-causal}. Counting both
causal-control arms, 12 predictions were registered, 4 became
unscoreable under their registered validity gates, and 5 of the 8
scored were correct, with every miss on the restoration side
(\cref{tab:pred-ledger}). The timing amendment (Step-6T, below)
registered 4 further predictions: 2 became unscoreable under its
margin-instrument gate and 1 of the 2 scored was correct.

\begin{table}[ht]
\caption{Causal-control cells: behavioral outcome (seeds
\textsc{recovered}-and-valid; seeds \textsc{displaced}) and final
contrast margin. Rescue doses in units of $\delta_{\mathrm{TS}}$;
kill rates are pronoun-flip probabilities. rule:prior is the
post-edit support ratio under the windowed counter (first pronoun
within 16 tokens of a cue): kill rows count the edited battery cue
class, rescue rows (*) the injected, battery-disjoint class
(App.~\ref{app:rescue}); the web-base entry is blank, the
focal cue class being unmeasurable by the frozen counter on the web
vocabulary. $\dag$: \textsc{intervention-invalid} (2/3 seeds fail
the registered KS control-family gate); its behavioral verdicts are
unscoreable, not failed.}
\label{tab:step6}
\centering\small
\begin{tabular}{@{}l r r ccc r@{}}
\toprule
arm & dose & rule:prior & rec/3 & died/3 & $\cm$-valid & mean $\cm$ \\
\midrule
TS base & --- & 7.91 & 3/3 & 0/3 & 3 & $+3.68$ \\
kill & 0.437 & 1.04 & 1/3 & 0/3 & 3 & $+0.27$ \\
kill & 0.645\textsuperscript{\dag} & 0.49 & 0/3 & 3/3 & 3 & $-0.38$ \\
kill & 1.0 & 0.02 & 0/3 & 3/3 & 2 & $-2.99$ \\
web base & --- & --- & 0/3 & 3/3 & 2 & $-0.52$ \\
rescue & 0.01 & 37.36* & 0/3 & 3/3 & 0 & $-0.30$ \\
rescue & 0.1 & 358* & 0/3 & 2/3 & 1 & $-0.18$ \\
rescue & 1.0 & 3{,}565* & 0/3 & 2/3 & 2 & $+0.16$ \\
rescue & 3.0 & 10{,}691* & 0/3 & 3/3 & 0 & $-0.28$ \\
\bottomrule
\end{tabular}

\end{table}

\paragraph{Design.} Support documents (\emph{girl}$\to$\emph{she} and \emph{boy}$\to$\emph{he},
symmetric) were injected into the collapsed web-packed corpus at
doses $\{0.01, 0.1, 1, 3\} \times \delta_{\mathrm{TS}}$, with an
equal token count of neutral documents removed, so every cell
matches the base corpus in size. A registered disjointness gate
verified that no injected frame, cue name, or margin-prompt
substring appears in any probe battery; the $n$-gram baseline
(App.~\ref{app:mech-alt}) confirms operationally that the injection
changes no battery-relevant surface statistic.

\paragraph{Verdicts (frozen evaluator; verifier agreed on all
keys).} R1 behavioral rescue at $d \geq 1$: \textsc{fail} (0/3
seeds \textsc{recovered} at every dose). R1c at $d{=}3$:
\textsc{fail}. R1 margin clause: \textsc{pass}; at $d{=}1$ the
cell-mean final $\cm$ is positive ($+0.16$), as is the margin in
each instrument-valid seed ($+0.12$, $+0.48$).
R2 (no rescue at trace dose): \textsc{pass}. R3 (dose monotonicity
$\rho \geq 0.8$): \textsc{fail}, $\rho_{\mathrm{rescue}} = +0.70$,
because the $3\times$ dose regresses (mean $\cm$ $-0.28$, below the
$1\times$ value). RS (specificity): \textsc{pass}. R1c/R2 margin
clauses: \textsc{instrument-invalid} (zero $\cm$-valid seeds at
$d{=}0.01$ and $d{=}3$). The registered falsifier fired on its
non-recovery clause ($d{=}1$ and $d{=}3$ both below two recovered
seeds); its blanket-\emph{she} artifact clause stayed silent
(artifact-flagged seeds per dose: $0/0/1/0$), so the failure is
genuine non-restoration and no corrupted success is hiding in it.

\paragraph{What moved and what did not.} Mean final $\cm$ by dose:
$-0.30$, $-0.18$, $+0.16$, $-0.28$. The instrument gate voids
verdict \emph{clauses}, not descriptive means: with $\cm$-valid seed
counts of $0/1/2/0$ across the four doses, the means at $0.01\times$
and $3\times$ average over gate-failing seeds only and carry no
instrument warranty; R3 used them as registered. The registered
margin clause at
$d{=}1$ passed (both instrument-valid seeds end positive) and
that was the only dose at which it did; the surrounding magnitudes
are small (all four dose means lie within $0.5$ nats of zero, with
two valid seeds at the passing dose and no registered uncertainty
estimate), so we read the margin movement as the registered verdict
states it and no further. Behavior never follows at any dose (OOD
conflict finals in the restoration cells stay at $0.08$--$0.33$,
\cref{sec:results-causal}). Mechanism and behavior, computed
independently throughout, dissociate here. Head-level
analyses (App.~\ref{app:mech2}) give the dissociation a finer
grain: at the $1\times$ dose the injection measurably rebuilds a
carrier head (the dominant last-layer head's zero-ablation cost
at the final checkpoint is $1.4$--$1.9$ nats in the
instrument-valid seeds, roughly double the un-injected web base,
with positive OV alignment), while the item-level bootstrap on
the battery (\cref{tab:mech-cis}) shows the behavioral drop still
excluding zero everywhere except the artifact-flagged run. The
dose buys a partial reassembly of the circuit while the behavior
stays collapsed.

\paragraph{Post-injection evidence ratios.}
The frozen battery-cue support counter cannot measure the injected
support, for a reason that is a design property rather than a bug:
the injected cue names are disjoint from every battery name (the
anti-teaching control above), and only $2$ of the $24$ are single
tokens in the web vocabulary, while the counter matches
single-token cues only. Verified directly: the frozen counter
returns identical class counts on the base corpus and all four
rescue corpora while total tokens grow by the injected amount. We
therefore re-ran the identical first-pronoun-within-16-tokens
window logic with the injected cue names matched as full token
sequences---defined, logged in the decisions record, and
launched before any output was read, but after all rescue verdicts
were known, so it is unit accounting, not a registered quantity.
Post-injection rule:prior ratios for the injected female-cue
class: $1.7$ (base web, the rescue names' natural background),
$37$, $358$, $3{,}565$, and $10{,}691$ across the
$\{0.01, 0.1, 1, 3\}\times\delta_{\mathrm{TS}}$ doses
(\cref{tab:step6}); the male-cue class behaves symmetrically. For
comparison, the battery girl-name ratio in the TinyStories cell
where the rule survives is $7.9$. Every rescue dose, including the
trace dose, places the injected construction's evidence ratio
above the level at which the rule survives elsewhere; the
restoration failure is not an artifact of doses that are
ratio-poor in context.

\paragraph{The timing test (Step-6T, registered amendment).} The
injection above is uniform-rate: its support is spread evenly
across the run, so the window before the collapse onset
(${\approx}2{,}800$ of $4{,}400$ steps; \cref{sec:results-mech})
receives only ${\approx}64\%$ of the nominal dose and the
remaining ${\approx}36\%$ arrives after the behavior has already
collapsed. Two adjacent literatures make timing a plausible
moderator: critical periods, where late removal of a deficit fails
to restore what early training would have built
\citep{achille2019critical}, and fact learning, where knowledge
injected late in pretraining corrupts rather than integrates
\citep{zucchet2025facts}. A registered amendment (with all tests,
gates, and a named outcome for every branch before launch)
therefore split the $1\times$ dose by
schedule: \texttt{rescue\_d100\_early} concentrates the full dose
into steps $0$--$2{,}400$, entirely before the collapse step, and
\texttt{rescue\_d100\_late} concentrates it into steps
$2{,}800$--$4{,}400$, entirely after it; three seeds each, scored by the
frozen evaluator (\texttt{scripts/eval\_step6t.py}).

The early window failed behaviorally (registered test T1:
\textsc{fail}). Zero of three seeds show a control-valid recovery;
final held-out conflict accuracies are $0.17/0.36/0.66$ (frozen
classes \textsc{never}/\textsc{unstable}/\textsc{unstable}), with
two of three ending below $0.5$. The third seed is the most
informative trajectory in the arm: it reaches conflict $1.00$
while the concentrated dose flows, then re-collapses after the
dose ends: the dose props the rule up, and withdrawal hands it
back to the prior. The late window failed exactly as the amendment
predicted (T2: \textsc{pass}; zero of three recoveries). The
mechanism-ordering tests returned no verdict: the margin
instrument gate (G4, at least two seeds with a valid $\cm$
readout) holds in the late cell but fails in the early cell ($1$
of $3$), so T1's margin clause and the cross-window ordering test
T3 are instrument-invalid under the registered rule, and the
falsifier, which requires the early gate, stayed silent for that
reason. The amendment's pre-named outcome for this branch is
``behavioral non-rescue, mechanism unscoreable.'' Scored hit rate:
$1$ of $2$ scorable, of $4$ registered.

The supported summary is therefore stronger than the uniform-rate
result alone: support removal is causally sufficient to destroy
the rule, and restoration at matched dose fails under a uniform
schedule, an early-concentrated schedule, and a late-concentrated
schedule alike. What the timing test leaves open is the
mechanism-level comparison (the margin instrument was valid in too
few early seeds to score) and any claim beyond this rule and this
scale.

\section{The Second-Rule (a/an) Kill Ladder in Full}
\label{app:ankill}

This appendix carries the full record of the second-rule kill
ladder summarized in \cref{sec:results-causal}: design, registered
predictions, per-seed outcomes, and the two registered misses.

\paragraph{Design.} The target rule is \emph{a}/\emph{an}
allomorphy, probed by two frozen battery families:
\texttt{det\_an\_choice} (article choice before a vowel-initial
noun; conflict items pit the rule against the corpus-dominant
\emph{a}) and \texttt{a\_an\_adjective} (the same rule before
vowel-initial adjectives). Base TinyStories contains $485{,}862$
rule-supporting events (\emph{an} before a vowel-initial word)
against $20{,}731$ counter-events (\emph{a} before a vowel-initial
word), an evidence ratio of $23.4$, and the rule survives at
ceiling in all three base seeds. The intervention flips
\emph{an}~$\to$~\emph{a} before vowel-initial words at rate $p$,
independently per eligible site, leaving every other token
untouched; each flip converts one supporting event into a
counter-event, so the post-edit evidence ratio is
$(1-p)\,S / (C + pS)$ for base counts $S, C$. Five doses
($p \in \{0.5, 0.667, 0.75, 0.9, 1.0\}$, spanning evidence ratios
$0.92$ down to $0$) were trained at three seeds each (42/43/44)
under the v1-reproduction configuration and scored with the frozen
\texttt{rvp31} battery and frozen classifier conventions, the
same machinery, unmodified, that scored every other cell in this
paper.

\paragraph{Registered predictions and verdicts.} Five blind
predictions (AK1--AK5) were registered in the decisions log
before any intervened corpus or run existed.
\textbf{AK1} (\textsc{pass}): final conflict accuracy decreases
monotonically in dose: the six cell means are strictly
decreasing. \textbf{AK2} (\textsc{pass}): the full flip kills the
rule: all three $p{=}1.0$ seeds end below $0.1$ (all at
$0.00$); with the rule never emerging at this dose (peaks $\leq
0.16$), the trained model behaves as if the construction does not
exist. \textbf{AK3} (\textsc{fail}): we predicted the boundary
would fall within a factor of two of the focal rule's, with
survival ($\geq 0.8$ in $\geq 2/3$ seeds) at the parity dose
$p{=}0.5$ and clear degradation by $p{=}0.75$. The degradation
clause held but the survival clause did not: $0/3$ seeds reach
$0.8$ at parity (finals $0.76/0.69/0.55$). Descriptively the two
boundaries are compatible (the focal rule crossed between
window ratios $0.487$ and $1.035$, and a/an destabilizes at
$0.92$), but the registered operationalization predicted
survival at $0.92$ and is scored as written. \textbf{AK4}
(\textsc{fail}): we predicted the agree-condition validity gate
would fail at the highest dose ($\geq 2/3$ seeds at $p{=}1.0$).
It never fired: agree accuracy stays $\geq 0.94$ in all $18$ runs.
This miss is benign for the ladder's interpretation (it is what
makes every behavioral verdict in \cref{tab:ankill} scoreable), but it is a miss, and it counts as one. \textbf{AK5}
(\textsc{pass}): the edit dissociates: pronoun-gender,
negation, and irregular-past family means stay within $0.05$ of
the base cell at every dose (observed range $0.98$--$1.00$).

Two scoring operationalizations were fixed at scoring time, before
verdicts were computed, and logged in the decisions record: AK4's
``gate fails'' reads as the frozen control-validity check failing
for \texttt{det\_an\_choice} in $\geq 2/3$ seeds at $p{=}1.0$, and
AK5's ``within noise'' reads as a $\pm 0.05$ tolerance on cell
means per family per dose. Scored verdicts live in
\texttt{runs/eval\_ankill.json}; the scoreboard rows in
\cref{tab:predictions} are generated from that artifact.

\section{Alternative Mechanisms Considered}
\label{app:mech-alt}

We list the alternative explanations a skeptic should consider, the
registered control that addresses each, and, where the control has
already run, its result.

\paragraph{``The rescue just biases the model toward `she'
globally.''} The injection is symmetric (\emph{girl}$\to$\emph{she} and
\emph{boy}$\to$\emph{he}); the agree condition acts as a per-seed validity gate; and
the registered rescue falsifier explicitly names the blanket-she
artifact (conflict recovery with a control-invalid agree condition in
$\ge 2/3$ seeds) as a failure mode, not a success. \emph{Outcome:}
moot in the strongest sense: the rescue falsifier did fire, but on
its \emph{non-recovery} clause; the blanket-she clause stayed silent
(artifact-flagged seeds per dose: $0/0/1/0$). There was no spurious
rescue to explain away, because there was no rescue.

\paragraph{``The kill removes data, so the effect is just less
data.''} The kill is token-count preserving: it flips single pronoun
token ids in place, and a registered gate (D4) asserts token-count
equality, with everything outside flipped positions byte-identical to
the source corpus.

\paragraph{``The kill damages the corpus generally.''} A registered
predicate (KS) requires four unrelated control families to remain
recovered-and-valid in every kill cell; if two or more fail in a cell,
that cell is \textsc{intervention-invalid} and yields no verdict.
\emph{Outcome:} the gate bound exactly once, at $p{=}0.645$, 2/3
seeds lost \texttt{det\_an\_choice} and \texttt{a\_an\_adjective}, so
that cell was voided as registered (K1 unscoreable, reported as such
in \cref{tab:step6}); the $p{=}0.437$ and $p{=}1.0$ cells passed the
gate, so the kill verdicts quoted in the main text rest only on
intervention-valid cells.

\paragraph{``An $n$-gram model would show the same dose-response, so
no capability is displaced.''} We trained a Stupid Backoff $n$-gram
baseline \citep{brants2007large} ($\alpha{=}0.4$; orders 2 and 5;
exact, query-restricted counts) on the \emph{exact token stream each
cell's model saw} (same shard order, same token budget) and
scored the frozen battery with the same forced-choice rule. The
interpretation rule was registered before any output was read.
Results (\cref{tab:ngram-pronoun,tab:ngram-full-0,tab:ngram-full-1,tab:ngram-full-2}):

\begin{itemize}
\item \emph{The baseline cannot exhibit the phase pattern at all}: on
held-out items its agree (control) condition fails in every relevant
cell ($0.33$ on the TinyStories base, $0.00$ on the web base at
$n{=}5$), so under the registered control-validity gate no cell-seed
of this model class is even classifiable. Surface co-occurrence does
not solve the held-out control condition, let alone the conflict
condition.
\item \emph{Kill side}: 5-gram conflict accuracy tracks the kill dose
($1.00 \to 0.17 \to 0.08 \to 0.00$), expected by construction,
since the kill edits pronoun tokens directly; this doubles as a
positive control on the intervention's surface-statistical strength.
Meanwhile all eleven non-target families are numerically identical
across kill doses (\cref{tab:ngram-full-0}), confirming surface-level
specificity of the kill.
\item \emph{Rescue side}: held-out battery scores are \emph{identical
to the un-injected base at every dose} (conflict $0.50$, agree
$0.00$; identical margins), across two orders of magnitude of dose.
The injected documents change no battery-relevant surface statistics, an operational confirmation of the registered disjointness gate,
and direct evidence that any behavioral rescue in the trained models
cannot be probe-surface leakage.
\item \emph{The displacement puzzle gets harder}: on the TinyStories
base corpus the 5-gram scores held-out conflict at $1.00$ (the
surface evidence for the rule is present throughout training), which is precisely what makes a trained model's loss of the rule on
stationary data a fact about learning dynamics rather than about the
data.
\end{itemize}

\begin{table}[t]
\caption{Stupid-backoff $n$-gram baseline ($\alpha=0.4$) on the frozen battery, pronoun family, heldout split, forced-choice argmax accuracy. The baseline is counted on the exact token stream each cell's model saw.}
\label{tab:ngram-pronoun}
\vskip 0.1in
\centering\small
\begin{tabular}{lcccc}
\toprule
 & \multicolumn{2}{c}{$n=2$} & \multicolumn{2}{c}{$n=5$} \\
\cmidrule(lr){2-3}\cmidrule(lr){4-5}
Cell & conflict & agree & conflict & agree \\
\midrule
TS base & 0.50 & 0.50 & 1.00 & 0.33 \\
kill p437 & 0.00 & 1.00 & 0.17 & 1.00 \\
kill p645 & 0.00 & 1.00 & 0.08 & 1.00 \\
kill p1000 & 0.00 & 1.00 & 0.00 & 1.00 \\
\midrule
web base & 0.00 & 1.00 & 0.50 & 0.00 \\
rescue d=0.01 & 0.00 & 1.00 & 0.50 & 0.00 \\
rescue d=0.1 & 0.00 & 1.00 & 0.50 & 0.00 \\
rescue d=1 & 0.00 & 1.00 & 0.50 & 0.00 \\
rescue d=3 & 0.00 & 1.00 & 0.50 & 0.00 \\
\midrule
TS D/N=5 & 0.50 & 0.50 & 1.00 & 0.08 \\
TS D/N=1.5 & 0.50 & 0.50 & 1.00 & 0.00 \\
web D/N=5 & 0.00 & 1.00 & 0.00 & 1.00 \\
web D/N=1.5 & 0.00 & 1.00 & 0.00 & 1.00 \\
\bottomrule
\end{tabular}
\end{table}

\begin{table*}[t]
\caption{$5$-gram baseline, all families, heldout argmax accuracy (conflict / agree): TinyStories base and kill cells.}
\label{tab:ngram-full-0}
\vskip 0.1in
\centering\small
\begin{tabular}{lcccc}
\toprule
Family & TS base & kill p437 & kill p645 & kill p1000 \\
\midrule
a\_an\_adjective & 1.00 / 1.00 & 1.00 / 1.00 & 1.00 / 1.00 & 1.00 / 1.00 \\
comparative\_er & 1.00 / 1.00 & 1.00 / 1.00 & 1.00 / 1.00 & 1.00 / 1.00 \\
det\_an\_choice & 1.00 / 1.00 & 1.00 / 1.00 & 1.00 / 1.00 & 1.00 / 1.00 \\
irregular\_past & 1.00 / 1.00 & 1.00 / 1.00 & 1.00 / 1.00 & 1.00 / 1.00 \\
irregular\_past\_v2 & 1.00 / 0.75 & 1.00 / 0.75 & 1.00 / 0.75 & 1.00 / 0.75 \\
modal\_agreement & 1.00 / 0.00 & 1.00 / 0.00 & 1.00 / 0.00 & 1.00 / 0.00 \\
modal\_agreement\_v2 & 1.00 / 0.06 & 1.00 / 0.06 & 1.00 / 0.06 & 1.00 / 0.06 \\
negation\_bare\_verb & 1.00 / 1.00 & 1.00 / 1.00 & 1.00 / 1.00 & 1.00 / 1.00 \\
negation\_bare\_verb\_v2 & 1.00 / 1.00 & 1.00 / 1.00 & 1.00 / 1.00 & 1.00 / 1.00 \\
plural\_was\_were & 1.00 / 1.00 & 1.00 / 1.00 & 1.00 / 1.00 & 1.00 / 1.00 \\
pronoun\_gender\_ref & 1.00 / 0.33 & 0.17 / 1.00 & 0.08 / 1.00 & 0.00 / 1.00 \\
reflexive\_gender & 0.29 / 0.79 & 0.29 / 0.79 & 0.29 / 0.79 & 0.29 / 0.79 \\
\bottomrule
\end{tabular}
\end{table*}

\begin{table*}[t]
\caption{$5$-gram baseline, all families, heldout argmax accuracy (conflict / agree): Web base and rescue cells.}
\label{tab:ngram-full-1}
\vskip 0.1in
\centering\small
\begin{tabular}{lccccc}
\toprule
Family & web base & rescue d=0.01 & rescue d=0.1 & rescue d=1 & rescue d=3 \\
\midrule
a\_an\_adjective & 0.88 / 1.00 & 0.88 / 1.00 & 0.88 / 1.00 & 0.88 / 1.00 & 0.88 / 1.00 \\
comparative\_er & 1.00 / 1.00 & 1.00 / 1.00 & 1.00 / 1.00 & 1.00 / 1.00 & 1.00 / 1.00 \\
det\_an\_choice & 0.99 / 1.00 & 0.99 / 1.00 & 0.99 / 1.00 & 0.98 / 1.00 & 0.98 / 1.00 \\
irregular\_past & 1.00 / 0.28 & 1.00 / 0.28 & 1.00 / 0.28 & 1.00 / 0.28 & 1.00 / 0.31 \\
irregular\_past\_v2 & 1.00 / 0.25 & 1.00 / 0.25 & 1.00 / 0.25 & 1.00 / 0.25 & 1.00 / 0.25 \\
modal\_agreement & 1.00 / 0.00 & 1.00 / 0.00 & 1.00 / 0.00 & 1.00 / 0.00 & 1.00 / 0.00 \\
modal\_agreement\_v2 & 1.00 / 0.04 & 1.00 / 0.04 & 1.00 / 0.04 & 1.00 / 0.04 & 1.00 / 0.04 \\
negation\_bare\_verb & 1.00 / 0.17 & 1.00 / 0.17 & 1.00 / 0.17 & 1.00 / 0.17 & 1.00 / 0.17 \\
negation\_bare\_verb\_v2 & 1.00 / 0.50 & 1.00 / 0.50 & 1.00 / 0.50 & 1.00 / 0.50 & 1.00 / 0.50 \\
plural\_was\_were & 1.00 / 0.94 & 1.00 / 0.94 & 1.00 / 0.94 & 1.00 / 0.94 & 1.00 / 0.94 \\
pronoun\_gender\_ref & 0.50 / 0.00 & 0.50 / 0.00 & 0.50 / 0.00 & 0.50 / 0.00 & 0.50 / 0.00 \\
reflexive\_gender & 0.08 / 1.00 & 0.08 / 1.00 & 0.08 / 1.00 & 0.08 / 1.00 & 0.08 / 1.00 \\
\bottomrule
\end{tabular}
\end{table*}

\begin{table*}[t]
\caption{$5$-gram baseline, all families, heldout argmax accuracy (conflict / agree): Data-budget cells.}
\label{tab:ngram-full-2}
\vskip 0.1in
\centering\small
\begin{tabular}{lcccc}
\toprule
Family & TS D/N=5 & TS D/N=1.5 & web D/N=5 & web D/N=1.5 \\
\midrule
a\_an\_adjective & 1.00 / 1.00 & 1.00 / 1.00 & 1.00 / 1.00 & 1.00 / 1.00 \\
comparative\_er & 1.00 / 1.00 & 1.00 / 1.00 & 1.00 / 1.00 & 1.00 / 1.00 \\
det\_an\_choice & 1.00 / 0.99 & 0.99 / 1.00 & 1.00 / 1.00 & 1.00 / 1.00 \\
irregular\_past & 1.00 / 0.97 & 1.00 / 0.94 & 1.00 / 0.08 & 1.00 / 0.03 \\
irregular\_past\_v2 & 1.00 / 0.75 & 1.00 / 0.75 & 1.00 / 0.25 & 1.00 / 0.17 \\
modal\_agreement & 1.00 / 0.12 & 1.00 / 0.06 & 1.00 / 0.00 & 1.00 / 0.06 \\
modal\_agreement\_v2 & 1.00 / 0.08 & 1.00 / 0.04 & 1.00 / 0.02 & 1.00 / 0.00 \\
negation\_bare\_verb & 1.00 / 1.00 & 1.00 / 1.00 & 1.00 / 0.08 & 1.00 / 0.04 \\
negation\_bare\_verb\_v2 & 1.00 / 0.88 & 1.00 / 0.75 & 1.00 / 0.38 & 1.00 / 0.38 \\
plural\_was\_were & 1.00 / 1.00 & 1.00 / 1.00 & 1.00 / 1.00 & 1.00 / 0.88 \\
pronoun\_gender\_ref & 1.00 / 0.08 & 1.00 / 0.00 & 0.00 / 1.00 & 0.00 / 1.00 \\
reflexive\_gender & 0.29 / 0.75 & 0.29 / 0.75 & 0.00 / 1.00 & 0.00 / 1.00 \\
\bottomrule
\end{tabular}
\end{table*}

\section{Second Mechanism Measure: Decomposition and Decodability}
\label{app:mech2}

\begin{figure*}[t]
\centering
\includegraphics[width=\textwidth]{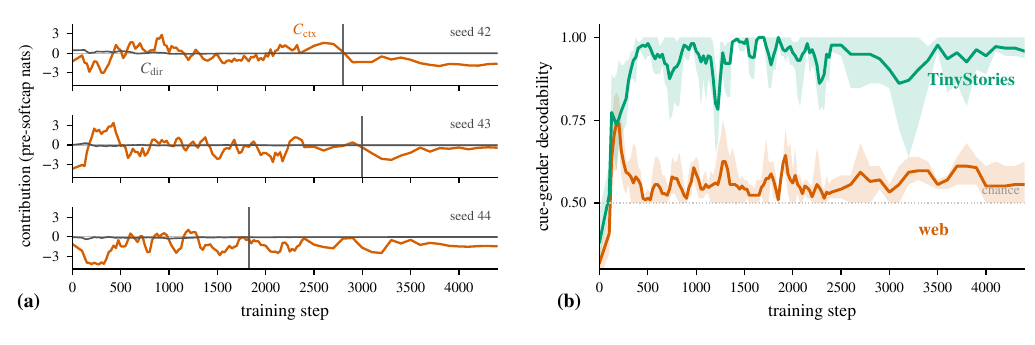}
\caption{The second mechanism measure. \textbf{(a)} Exact
decomposition of the pre-softcap \emph{she}--\emph{he} gap on the
conflict prefixes, per web seed: the contextual contribution
$C_{\mathrm{ctx}}$ (vermillion) carries the margin and ends
negative, the direct embedding-path contribution $C_{\mathrm{dir}}$
(grey) stays near zero throughout; vertical line = behavioral
collapse onset. The measure requires no peak-height gate, so all
three seeds are shown at full strength, including seed 44, which
the $\cm$ instrument rejects. \textbf{(b)} Cross-frame
nearest-class-mean decodability of the cue's gender in the
final-layer residual at the prediction site (seed mean, min--max
band): after an early transient, marginal on web through the
rule's lifetime and near chance by the end: the cue's gender is
no longer transported to where the pronoun is predicted.}
\label{fig:mech2}
\end{figure*}

\textbf{Status.} This measure was defined and committed
(\texttt{logs/DECISIONS.md}) before any checkpoint was scored with it, but after the registered $\cm$ results were known. It is corroboration for the displacement
mechanism, never confirmatory evidence, and no registered verdict
depends on it. It is computed by offline rescoring of the same
frozen checkpoints and the same frozen battery slices as the $\cm$
instrument (conflict/heldout prefixes as the girl-cue set, agree/heldout
as the boy-cue set; identical frames), so no new prompts were
authored after results were known.

\textbf{Direct-logit decomposition.} Following the residual-stream
view of \citet{elhage2021mathematical}, the final residual state is an
exact sum of the embedding stream and every block's attention and
MLP increments,
$x_{\mathrm{final}} = \mathrm{rmsnorm}(\mathrm{wte}) + \sum_i a_i +
\sum_i m_i$, and the pre-softcap logit gap is
$x_{\mathrm{final}} \cdot (w_{she} - w_{he}) / \mathrm{rms}(
x_{\mathrm{final}})$, so each component's contribution is exact
given the realized rms (reconstruction error $<10^{-3}$ asserted at
every checkpoint; the replicated forward path reproduces the
registered $\cm$ values exactly at all 114 checkpoints per run).
The model caps logits with a $\tanh$ soft-cap as in
\citet{gemmateam2024gemma2}, hence ``pre-softcap.''
$C_{\mathrm{dir}}$ is the embedding-stream term;
$C_{\mathrm{ctx}}$ is the sum of all attention and MLP terms. The
softcap is monotone elementwise, so the pre- and post-cap gaps
agree in sign; pre-cap amplitudes run larger than $\cm$. Because no
peak-height validity gate applies, $C_{\mathrm{ctx}}$ is defined on
all seeds. Result (\cref{fig:mech2}a): in all three web seeds
$C_{\mathrm{dir}}$ is pinned near zero from early training onward
(final values $-0.01$ to $-0.05$ nats) while $C_{\mathrm{ctx}}$
rises with the rule and ends negative ($-1.6/-0.4/-1.5$ smoothed).
The surviving TinyStories runs show the same structure with the
opposite sign: $C_{\mathrm{dir}}$ stays below $0.1$ nats while
$C_{\mathrm{ctx}}$ ends at $+8$ to $+14$. The margin---and its
collapse---lives entirely in the contextual pathway; surviving
and collapsed runs differ in the sign of the contextual term, and
the path itself never switches. At layer grain the
attention and MLP terms are large and partially cancelling, and we
claim nothing there; the head-grain analyses below, registered
separately, give a finer-grained view at the prediction position.

\textbf{Cue-gender decodability.} At each checkpoint we take the
rms-normalized residual at the final (prediction) position after
each block, for girl-cue and boy-cue prompts, and classify cue
gender with a cross-frame nearest-class-mean rule: class means are
fit on one syntactic frame's prompts and tested on the other's, in
both directions. The classifier has no trained parameters and no
hyperparameters, and is unbiased at chance $0.5$ under no signal. A
leave-one-out variant ran first and was replaced when its first
partial outputs sat far below $0.5$; synthetic-noise tests confirmed
that variant is chance-biased downward under no signal, the
replacement unbiased, and the swap was made and logged on bias
grounds before any complete run had been scored. The embedding
layer scores $0$ by construction (same-frame prompts share their
final token, so all distances tie and ties score as errors) and is
excluded. Result (\cref{fig:mech2}b): on web, apart from a brief
transient that precedes the rule's behavioral emergence (smoothed
seed-mean peak $0.74$ at step $175$), final-layer decodability
stays within $0.50$--$0.64$ through the rule's behavioral lifetime
and ends at $0.56$; in the surviving TinyStories runs the same
instrument runs near ceiling (smoothed seed mean $0.94$ from step
$500$ on, $0.96$ at the end).
This is the representation-level face of displacement: after
collapse the network no longer delivers the cue's gender to the
prediction site, against the alternative that an intact rule is
merely outvoted at the logit layer, which would leave the signal
decodable. The instrument is a lower bound (nearest-class-mean
can miss low-variance directions), so the evidence is the
web-vs-TinyStories contrast under the identical instrument; the
absolute level carries no weight.

\textbf{Head-level analyses (same status).} A second battery of
analyses (per-head attribution and ablation, OV alignment
\citep{elhage2021mathematical},
same-run activation patching \citep{vig2020causal,meng2022locating},
direction specificity, and item-level
bootstrap CIs) was registered the same way (committed to the
decisions log before any checkpoint was scored with
it, after all registered verdicts were known) and carries the same
status: descriptive corroboration over the same frozen checkpoints,
prompts, and scorer; no registered verdict depends on any of it. It
covers the base cells and the causal-control cells at full kill and
the two highest rescue doses
(\cref{tab:mech-heads,tab:mech-patch,tab:mech-cis}).

\textbf{Per-head attribution and ablation.} The attention term of
the decomposition splits exactly over heads (each layer's
value-embedding mixture is attributed to the head that reads it;
head terms sum to the layer term, asserted to $<10^{-3}$ at every
checkpoint). In the surviving TinyStories runs the margin rides a
single last-layer head: one head carries $0.75$--$0.90$ of the
total head-level $|$attribution$|$ at the $\cm$ peak and
$0.75$--$0.85$ at the final checkpoint, and zero-ablating that one
head at all positions deletes essentially the entire post-softcap
margin (e.g.\ $-20.3$ of $+20.5$ nats at the strongest peak;
$-2.7$ to $-3.8$ of the $+2.4$ to $+4.8$ final margins). Which head
plays the role differs by seed (L3H0 or L3H1): the structure
repeats across seeds while the head's identity varies, consistent
with the
seed-robust-displacement framing of \cref{sec:results-mech}. In the
collapsed web runs no such carrier exists: top-head share
$0.27$--$0.52$, the identity unstable between peak and final and
across seeds, single-head ablation effects fractional
(\cref{tab:mech-heads}).

\begin{table}[h]
\caption{Dominant last-position head per run (largest $|$attribution$|$ to the pre-softcap \emph{she}--\emph{he} gap), at the run's own $\cm$-peak checkpoint and at the final checkpoint. share = that head's fraction of total head-level $|$attribution$|$; $\Delta\cm_{\mathrm{abl}}$ = change in post-softcap $\cm$ when the head's output is zeroed at all positions; cos = cosine of the head's mean OV output on feminine cue tokens with $w_{she}-w_{he}$. Descriptive, post-hoc (App.~\ref{app:mech2}).}
\label{tab:mech-heads}
\vskip 0.1in
\centering\small
\begin{tabular}{llrlrrrrlrrr}
\toprule
 & & \multicolumn{5}{c}{at $\cm$ peak} & \multicolumn{5}{c}{at final} \\
\cmidrule(lr){3-7}\cmidrule(lr){8-12}
Cell & seed & $\cm$ & head & share & $\Delta\cm_{\mathrm{abl}}$ & cos & $\cm$ & head & share & $\Delta\cm_{\mathrm{abl}}$ & cos \\
\midrule
TS base & 42 & $+20.53$ & L3H0 & 0.85 & $-20.28$ & $+0.49$ & $+3.83$ & L3H0 & 0.75 & $-3.79$ & $+0.54$ \\
 & 43 & $+22.32$ & L3H1 & 0.90 & $-22.29$ & $+0.69$ & $+2.42$ & L3H1 & 0.85 & $-2.74$ & $+0.59$ \\
 & 44 & $+13.77$ & L3H1 & 0.77 & $-7.72$ & $+0.45$ & $+4.79$ & L3H1 & 0.81 & $-3.63$ & $+0.50$ \\
\midrule
web base & 42 & $+1.46$ & L3H1 & 0.52 & $-1.98$ & $+0.09$ & $-0.75$ & L2H1 & 0.39 & $-0.35$ & $+0.04$ \\
 & 43 & $+1.89$ & L1H0 & 0.31 & $-0.08$ & $+0.39$ & $-0.27$ & L3H1 & 0.35 & $-1.01$ & $+0.41$ \\
 & 44 & $+1.08$ & L1H0 & 0.33 & $-0.77$ & $+0.14$ & $-0.64$ & L3H0 & 0.27 & $-0.63$ & $+0.38$ \\
\midrule
kill $p{=}1$ & 42 & $+0.90$ & L3H0 & 0.40 & $+0.64$ & $+0.07$ & $-4.74$ & L3H0 & 0.58 & $-1.83$ & $+0.20$ \\
 & 43 & $+1.40$ & L3H0 & 0.26 & $-2.93$ & $+0.13$ & $-1.04$ & L3H1 & 0.56 & $-2.62$ & $+0.20$ \\
 & 44 & $+2.70$ & L3H0 & 0.51 & $-2.32$ & $+0.12$ & $-3.80$ & L3H1 & 0.64 & $+3.41$ & $-0.32$ \\
\midrule
rescue $d{=}1$ & 42 & $+1.10$ & L3H0 & 0.37 & $-0.13$ & $+0.01$ & $-0.22$ & L3H1 & 0.38 & $-0.54$ & $+0.34$ \\
 & 43 & $+1.51$ & L3H0 & 0.56 & $-1.23$ & $+0.20$ & $+0.16$ & L3H0 & 0.55 & $-1.86$ & $+0.18$ \\
 & 44 & $+1.18$ & L3H0 & 0.41 & $-0.89$ & $+0.44$ & $+0.44$ & L3H0 & 0.51 & $-1.38$ & $+0.42$ \\
\midrule
rescue $d{=}3$ & 42 & $+0.70$ & L3H0 & 0.29 & $-0.42$ & $-0.01$ & $-0.01$ & L3H0 & 0.43 & $-0.12$ & $-0.00$ \\
 & 43 & $+0.69$ & L3H0 & 0.38 & $-1.18$ & $+0.24$ & $-0.49$ & L3H0 & 0.68 & $-2.04$ & $+0.25$ \\
 & 44 & $+0.64$ & L3H1 & 0.54 & $-0.79$ & $+0.08$ & $-0.28$ & L3H1 & 0.41 & $-0.50$ & $+0.12$ \\
\bottomrule
\end{tabular}
\end{table}

\textbf{OV alignment.} The cosine between the dominant head's mean
OV output on the frozen feminine-cue tokens and the readout
direction $w_{she}-w_{he}$ is $0.45$--$0.69$ in the surviving runs
at both checkpoints; in the collapsed web runs the (smaller,
unstable) dominant head's alignment spans $0.04$--$0.41$. The full
kill produces the one sign reversal in the sweep: at the final
checkpoint of one $p{=}1$ seed the dominant head is
\emph{anti}-aligned (cosine $-0.32$, attribution $-4.7$ nats): under fully reversed evidence the carrier position ends up writing
the masculine direction, a displacement by outright reversal.

\textbf{Same-run activation patching.} Thirteen component outputs
(embedding stream, eight heads, four MLP increments) are cached on
the frozen cue prompts at one of the run's own checkpoints and
swapped, one at a time, into the other checkpoint's forward pass
(\cref{tab:mech-patch}). In every collapsed web run the two
directions are asymmetric: no single peak component patched into
the final model recovers more than $0.38$ of the peak-minus-final
margin gap, while a single late-MLP component of the final model
patched into the peak model reproduces $0.5$--$1.6\times$ the full
collapse. Descriptively, one displaced component suffices to
destroy the margin and no single preserved component suffices to
restore it, the same asymmetry the corpus-level interventions
show behaviorally (\cref{sec:results-causal}), visible inside a
single run's own trajectory. The surviving TinyStories runs, where
the final checkpoint still carries the rule and the gap merely
shrinks, show no such consistent pattern;
recovery there is distributed and overshoots occur in both
directions.

\begin{table}[h]
\caption{Single-component activation patching between each run's own $\cm$-peak and final checkpoints (cached component outputs on the frozen feminine-cue prompts, swapped at all positions). Each entry is the component with the largest recovery fraction $(\cm_{\mathrm{patch}}-\cm_{\mathrm{dst}})/(\cm_{\mathrm{src}}-\cm_{\mathrm{dst}})$; fractions above $1$ overshoot the source value. Descriptive, post-hoc (App.~\ref{app:mech2}).}
\label{tab:mech-patch}
\vskip 0.1in
\centering\small
\begin{tabular}{lllrlr}
\toprule
 & & \multicolumn{2}{c}{peak $\to$ final} & \multicolumn{2}{c}{final $\to$ peak} \\
\cmidrule(lr){3-4}\cmidrule(lr){5-6}
Cell & seed & component & rec. & component & rec. \\
\midrule
TS base & 42 & MLP 2 & $+0.63$ & MLP 3 & $+1.12$ \\
 & 43 & MLP 1 & $+0.35$ & MLP 3 & $+1.18$ \\
 & 44 & MLP 0 & $+1.72$ & MLP 0 & $+1.53$ \\
\midrule
web base & 42 & MLP 3 & $+0.34$ & MLP 3 & $+1.63$ \\
 & 43 & MLP 1 & $+0.18$ & MLP 3 & $+0.82$ \\
 & 44 & MLP 3 & $+0.38$ & MLP 0 & $+0.52$ \\
\midrule
kill $p{=}1$ & 42 & MLP 2 & $+0.84$ & MLP 2 & $+1.67$ \\
 & 43 & MLP 2 & $+2.89$ & MLP 3 & $+0.67$ \\
 & 44 & head L3H1 & $+0.43$ & MLP 3 & $+0.68$ \\
\midrule
rescue $d{=}1$ & 42 & MLP 2 & $+0.36$ & MLP 2 & $+1.56$ \\
 & 43 & MLP 3 & $+1.48$ & MLP 2 & $+0.66$ \\
 & 44 & head L1H0 & $+0.25$ & MLP 2 & $+0.39$ \\
\midrule
rescue $d{=}3$ & 42 & MLP 0 & $+1.15$ & MLP 3 & $+4.00$ \\
 & 43 & MLP 3 & $+0.51$ & MLP 3 & $+0.74$ \\
 & 44 & head L0H0 & $+0.26$ & MLP 3 & $+2.15$ \\
\bottomrule
\end{tabular}
\end{table}

\textbf{Direction specificity.} At each post-block depth a
class-mean feminine-minus-masculine direction is fit on the frozen
cue prompts and projected out of the residual stream at all
positions. In the TinyStories runs at peak, removing the single
final-layer direction costs most of the margin ($\Delta\cm$ $-9.4$
to $-16.1$ nats against margins of $+13.8$ to $+22.3$), while the
mean absolute change it induces across the eleven non-target
families' frozen margins is $9$--$18\times$ smaller
($0.76$--$1.25$ nats) and the validation bits-per-byte cost is
$+0.07$ to $+0.11$: the direction is specific to the rule and
carries little of the general computation. In the collapsed web
runs no fitted direction at any depth moves $\cm$ by more than
$0.22$ nats, consistent with the decodability result: after
displacement there is no coherent gender direction left to remove.

\textbf{Item-level uncertainty.} The frozen probe log stores
aggregates, so the peak and final checkpoints of each run were
rescored at item level with the frozen scorer and given
template-stratified bootstrap CIs, with the same item resample
applied to both checkpoints so the peak-minus-final drop respects
the item pairing (\cref{tab:mech-cis}). Every web-base, kill
$p{=}1$ (pronoun), and rescue drop excludes zero at $95\%$, with
one exception: the pronoun drop of the registered artifact-flagged
rescue run. The TinyStories drops
are zero to within resolution. The full kill also shows item-level
specificity within the gender families themselves: the pronoun
family it edits drops by $0.50$--$1.00$ while the untouched
reflexive family's drop is exactly zero in all three seeds.

\begin{table}[h]
\caption{Item-level bootstrap on the displaced gender families (heldout conflict split; pronoun $n{=}12$ items, reflexive $n{=}24$): final-checkpoint accuracy, and the peak-minus-final drop with its 95\% CI from 10{,}000 template-stratified resamples applied jointly to the peak and final checkpoints (the pairing is preserved, so the drop CI accounts for item-level correlation). The peak checkpoint per family is the probe-log argmax snapped to the nearest stored checkpoint. The rescue $d{=}1$ seed-44 run is the registered blanket-\emph{she} artifact-flagged run (App.~\ref{app:rescue}); its near-zero pronoun drop fails the agree-condition control. Descriptive, post-hoc.}
\label{tab:mech-cis}
\vskip 0.1in
\centering\small
\begin{tabular}{llrlrl}
\toprule
 & & \multicolumn{2}{c}{pronoun\_gender\_ref} & \multicolumn{2}{c}{reflexive\_gender} \\
\cmidrule(lr){3-4}\cmidrule(lr){5-6}
Cell & seed & final & drop [95\% CI] & final & drop [95\% CI] \\
\midrule
TS base & 42 & $1.00$ & $+0.00$ \,{\scriptsize$[+0.00,+0.00]$} & $1.00$ & $+0.00$ \,{\scriptsize$[+0.00,+0.00]$} \\
 & 43 & $1.00$ & $+0.00$ \,{\scriptsize$[+0.00,+0.00]$} & $1.00$ & $+0.00$ \,{\scriptsize$[+0.00,+0.00]$} \\
 & 44 & $1.00$ & $+0.00$ \,{\scriptsize$[+0.00,+0.00]$} & $0.96$ & $+0.04$ \,{\scriptsize$[+0.00,+0.12]$} \\
\midrule
web base & 42 & $0.00$ & $+1.00$ \,{\scriptsize$[+1.00,+1.00]$} & $0.38$ & $+0.62$ \,{\scriptsize$[+0.46,+0.79]$} \\
 & 43 & $0.17$ & $+0.83$ \,{\scriptsize$[+0.67,+1.00]$} & $0.08$ & $+0.83$ \,{\scriptsize$[+0.67,+0.96]$} \\
 & 44 & $0.17$ & $+0.83$ \,{\scriptsize$[+0.67,+1.00]$} & $0.17$ & $+0.75$ \,{\scriptsize$[+0.62,+0.88]$} \\
\midrule
kill $p{=}1$ & 42 & $0.00$ & $+0.75$ \,{\scriptsize$[+0.58,+0.92]$} & $1.00$ & $+0.00$ \,{\scriptsize$[+0.00,+0.00]$} \\
 & 43 & $0.50$ & $+0.50$ \,{\scriptsize$[+0.50,+0.50]$} & $1.00$ & $+0.00$ \,{\scriptsize$[+0.00,+0.00]$} \\
 & 44 & $0.00$ & $+1.00$ \,{\scriptsize$[+1.00,+1.00]$} & $1.00$ & $+0.00$ \,{\scriptsize$[+0.00,+0.00]$} \\
\midrule
rescue $d{=}1$ & 42 & $0.17$ & $+0.83$ \,{\scriptsize$[+0.67,+1.00]$} & $0.25$ & $+0.75$ \,{\scriptsize$[+0.62,+0.88]$} \\
 & 43 & $0.58$ & $+0.42$ \,{\scriptsize$[+0.25,+0.50]$} & $0.08$ & $+0.88$ \,{\scriptsize$[+0.75,+1.00]$} \\
 & 44 & $0.92$ & $+0.08$ \,{\scriptsize$[+0.00,+0.25]$} & $0.46$ & $+0.50$ \,{\scriptsize$[+0.38,+0.62]$} \\
\midrule
rescue $d{=}3$ & 42 & $0.42$ & $+0.58$ \,{\scriptsize$[+0.33,+0.83]$} & $0.08$ & $+0.92$ \,{\scriptsize$[+0.83,+1.00]$} \\
 & 43 & $0.00$ & $+0.92$ \,{\scriptsize$[+0.75,+1.00]$} & $0.33$ & $+0.67$ \,{\scriptsize$[+0.54,+0.79]$} \\
 & 44 & $0.25$ & $+0.75$ \,{\scriptsize$[+0.50,+1.00]$} & $0.17$ & $+0.83$ \,{\scriptsize$[+0.71,+0.96]$} \\
\bottomrule
\end{tabular}
\end{table}

\section{Extended Related Work}
\label{app:related-ext}

\paragraph{Grokking and its variants.} Grokking
\citep{power2022grokking} is delayed generalization: train accuracy
saturates long before test accuracy jumps. Mechanistic accounts
explain the jump as circuit formation \citep{nanda2023progress} or as
a competition between memorizing and generalizing circuits under
efficiency pressure \citep{varma2023explaining};
\citet{liu2022omnigrok} show the phenomenon extends beyond modular
arithmetic, \citet{murty2023hierarchical} observe it for
hierarchical structure in transformers trained on natural language,
and \citet{liu2022understanding} draw phase diagrams of grokking
regimes over data size and hyperparameters, differing from ours in
axes (training-set size and model size against our frequency
statistic of natural text and $\dn$) and in charting a rise rather
than a within-run reversal. Varma et al.'s ``ungrokking'' (generalization receding
when the data budget shrinks below a critical size) is the closest
dynamical relative of our subject, but it is induced by changing the
dataset between training runs. The phenomenon we study has the
opposite sign and a different trigger: a capability is acquired
\emph{early}, then lost \emph{within a single run on stationary
data}, while the evidence for the rule remains in every epoch (and,
as the $n$-gram baseline in App.~\ref{app:mech-alt} shows, remains
recoverable from surface statistics throughout). Circuit-competition
accounts predict which circuit wins as a function of efficiency; our
phase diagram makes the analogous prediction as a function of two
measurable data quantities, $\dn$ and support frequency, and tests it
causally.

\paragraph{Forgetting.} Catastrophic forgetting
\citep{kirkpatrick2017ewc} and example forgetting
\citep{toneva2019forgetting} concern interference under distribution
shift or sample-level dynamics under SGD; \citet{tirumala2022memorization}
and \citet{jagielski2023measuring} study memorization and its decay in
LM pretraining, where earlier-seen examples are forgotten as training
continues on \emph{new} data. In all of these, forgetting is driven by
the disappearance or dilution of the supporting data. Closer to our
setting, \citet{chang2024characterizing} document that individual
tokens are forgotten during ordinary pretraining on a fixed
distribution, at rates that depend on token frequency, so
frequency-dependent forgetting on stationary data is not by itself
new. What is new here is the level and the structure: a rule-level
capability, verified on held-out conflict items against a surface
prior, that is acquired, exercised, and then displaced; a phase
diagram in $(\support, \dn)$ for whether that happens; and a
mechanism-plus-intervention account of why. The continued presence
of the rule's evidence through the loss is what makes displacement
(a competing prior crowding out a learned rule) rather than erasure
the natural mechanistic hypothesis.

\paragraph{Frequency and grammar learning in LMs.} The
psycholinguistically flavored literature anticipates parts of our
frequency axis. \citet{wei2021frequency} causally vary the
pretraining frequency of subject--verb agreement evidence and show
that rule behavior tracks absolute and relative frequency;
\citet{chang2022word} relate acquisition age to frequency; at
scale, factual-recall accuracy tracks the frequency of supporting
documents at final checkpoints \citep{kandpal2023longtail}, where
we chart within-run fate; \citet{choshen2022grammar} document
shared, sometimes non-monotonic phenomenon-level learning curves
across LMs; and \citet{warstadt2020learning} show that the shift
from surface heuristics to linguistic generalization is
data-dependent. The
filtered-corpus line is methodologically closest to our kill:
\citet{patil2024fict} remove a construction's direct evidence at
scale and find BabyLM-class models still learn it from indirect
evidence, and \citet{misra2024aanns} reach the same conclusion for
the rare article-adjective-numeral-noun construction. Their
ablations and our intervention answer different questions. Filtered
corpora withhold positive evidence; our edit flips it in place, so
that at full dose every supporting event becomes a counter-example
while token counts and all other statistics are held fixed. The
near-parity dose $p{=}0.437$ (the rung closest to a pure
ablation, thinning support without net contradiction) does not
kill the rule ($0/3$ seeds \textsc{displaced}), the outcome the
filtered-corpus results would
suggest; death arrives only as the edit approaches contradiction.
The two bodies of evidence are therefore consistent, and jointly
they narrow the claim: indirect evidence can carry a rule when
direct support is withheld, but it cannot defend one against
reversed support. The flip edit itself is counterfactual data
augmentation in the tradition of gender-swap debiasing
\citep{zhao2018gender, lu2020gender}; what is new is its use as a
registered, dose-graded causal probe under blind directional
predictions.

\paragraph{Critical learning periods.} \citet{achille2019critical}
showed that deficits imposed early in training cause permanent
capability loss in deep networks. We impose no deficit:
the corpus is stationary and the schedule fixed; the ``sensitive
period'' structure we observe arises from the data distribution
itself. The kill intervention can be read as a converse experiment:
rather than depriving the network of input during a window, it
removes the statistical support for one rule uniformly in time and
asks whether the acquisition-then-loss trajectory shifts as the
theory predicts.

\paragraph{Emergence and phase transitions in LMs.}
Emergent-abilities claims \citep{wei2022emergent} and their
metric-artifact critique \citep{schaeffer2023mirage} concern
capability appearance as a function of \emph{scale}. Our axis is
training time at fixed scale, where sudden structural change is well
documented: induction heads form in an abrupt window
\citep{olsson2022induction}, syntax acquisition in MLMs shows phase
transitions and simplicity bias \citep{chen2024sudden},
emergent in-context learning can itself be transient, fading as
in-weights solutions take over \citep{singh2023transient}, and the
developmental-interpretability program \citep{hoogland2024developmental,
wang2024refined} gives quantitative tools (local learning
coefficients) for locating such transitions. \citet{chen2024sudden}
is the closest precedent (capabilities moving non-monotonically
during pretraining) but documents the rise; our subject is the
fall, its predictability, and its causal control. The
forced-choice minimal-pair methodology descends from BLiMP
\citep{warstadt2020blimp}; our battery differs in pitting a
\emph{rule} against a \emph{prior} (conflict items) while measuring
the same rule where rule and prior agree (control items), which is
what lets a trajectory distinguish rule loss from global drift.
Crosscoders \citep{lindsey2024crosscoders} motivated our model-diffing
framing of displacement, though the confirmatory mechanism metric in
this paper is the simpler contrast-margin instrument
(App.~\ref{app:technical}).

\paragraph{Data budgets and mixtures.} Compute-optimal scaling
\citep{hoffmann2022chinchilla} fixes the token budget axis of our
grid; \citet{muennighoff2023data} characterize the data-constrained
regime (our $\dn$ axis is the reciprocal lens on the same quantity);
and mixture-optimization work \citep{xie2023doremi,diao2025climb}
treats domain weights as the controllable knob. Our causal knob is
finer-grained: the within-domain frequency of support for a single
linguistic rule, moved in both directions at matched token counts.
Corpora: TinyStories \citep{eldan2023tinystories} and a web mixture
derived from ClimbMix \citep{diao2025climb}. The public-suite
validation uses Pythia \citep{biderman2023pythia} and OLMo
\citep{groeneveld2024olmo} training-revision checkpoints.

\end{document}